\RequirePackage{snapshot}

\documentclass[10pt,twocolumn]{article}
\usepackage{authblk}

\usepackage[top=1in, bottom=1in, left=0.7in, right=0.7in]{geometry}

\usepackage{color}
\usepackage{times}
\usepackage{epsfig}
\usepackage{amsmath}
\usepackage{amssymb}
\usepackage{booktabs}            
\usepackage{multirow}
\usepackage[pagebackref=true,breaklinks=true,letterpaper=true,colorlinks,bookmarks=false]{hyperref}
\usepackage{nicefrac}

\usepackage{footnote} 

\newcommand{\bx}{\mathbf{x}}

\definecolor{tbLcolor}{rgb}{0.50, 0, 0}
\definecolor{tbScolor}{rgb}{0.75, 0, 0}

\definecolor{tb1color}{rgb}{0.40, 0, 0}
\definecolor{tb2color}{rgb}{0.55, 0, 0}
\definecolor{tb3color}{rgb}{0.65, 0, 0}
\definecolor{tb4color}{rgb}{0.75, 0, 0}
\definecolor{tb5color}{rgb}{0.85, 0, 0}
\newcommand{\tba}[1]{\textbf{\color{tbScolor}#1}}
\newcommand{\tbb}[1]{\textbf{\color{tbLcolor}#1}}
\newcommand{\tbc}[1]{\textbf{\color{tbLcolor}#1}}
\newcommand{\tbd}[1]{\textbf{\color{tbLcolor}#1}}

\newcommand{\tb}[1]{#1} 

\newcommand*{\doi}[1]{}

\graphicspath{{./figs/}}

\begin{document}

\title{Comparing Feature Detectors:\\ A bias in the repeatability criteria,\\ and how to correct it}

\author{Ives Rey-Otero$^{\dagger}$ \qquad Mauricio Delbracio $^{\star \dagger}$ \qquad Jean-Michel Morel$^{\dagger}$}
\affil{$^{\dagger}$CMLA, ENS-Cachan, France  \qquad $^{\star}$ECE, Duke University, USA}



%
%
%


\maketitle

\begin{abstract}
Most computer vision application rely on algorithms finding local
correspondences between different images.
These algorithms detect and compare stable local invariant descriptors centered at scale-invariant keypoints.
Because of the importance of the problem,
new keypoint detectors and descriptors are constantly being proposed, each one
claiming to perform better (or to be complementary) to the preceding ones. This
raises the question of a fair comparison between very diverse methods.
This evaluation has been mainly based on a repeatability criterion of the keypoints
under a series of image perturbations (blur, illumination, noise, rotations, homotheties,
homographies, etc).
In this paper, we argue that the classic repeatability
criterion is biased towards algorithms producing redundant overlapped
detections.
To compensate this bias, we propose a
variant of the repeatability rate taking into account the descriptors
overlap.
We apply this variant to revisit the popular benchmark by Mikolajczyk et al.~\cite{Miko2005}, on classic and new feature detectors.
Experimental evidence shows that the hierarchy of these feature detectors is severely
disrupted by the amended comparator.
\end{abstract}




%




\section{Introduction}
\label{sec:intro}

Local stable features are the cornerstone of many image processing and computer
vision  applications such as image
registration~\cite{hartley2003multiple,snavely2006photo}, camera
calibration~\cite{grompone2010towards}, image
stitching~\cite{haro2012photographing},  3\textsc{d}
reconstruction~\cite{agarwal2011building}, object
recognition~\cite{grimson1990object,fergus2003object,bay2006interactive,zhang2007local}
or visual
tracking~\cite{reid1979algorithm,zhou2009object}.
The seminal SIFT method introduced by D.~Lowe in 1999~\cite{Lowe1999,Lowe2004}
sparked an explosion of local keypoints detector/descriptors seeking
discrimination and invariance to a specific group of image
transformations~\cite{tuytelaars2008local}.
While deep neural networks \cite{hinton2012nips,bengio2013representation} have recently re-emerged
giving  state-of-the-art  performance in many computer vision activities, a
wide range of image processing tasks still rely on the extraction and
description of stable invariant keypoints.

Ideally, one would like to detect keypoints that are stable to image noise,
illumination changes, and geometric transforms such as scale changes,
affinities, homographies, perspective changes, or non-rigid deformations.
Complementarily, the detected points should be well distributed throughout the
entire image to extract information from all image regions and from boundary
features of all kinds (e.g., textures, corners, blobs).
Hence, there is a variety of detectors/descriptors built on different
principles and having different requirements.
While the SIFT method and its similar
competitors~\cite{bay2006surf,Miko2005,sifer} detect blob like structure in a
multi-scale image decomposition, other
approaches~\cite{brown2005multi,Miko2005,rosten2006machine,forstner2009detecting,
rosten2010faster,leutenegger2011brisk} explicitly detect corners or junctions
at different scales.
As opposed to interest point detectors, interest region detectors
\cite{tuytelaars1999content,tuytelaars2000wide,kadir2004affine,cao2005extracting}  extract the
invariant salient regions of an image based on its topographic map.
To fairly compare the very different feature detectors it is fundamental to have a rigorous evaluation protocol.

Introduced for the assessment of corner detectors
\cite{haralick1993computer} and later reformulated to evaluate
scale/affine-invariant keypoint
detectors~\cite{schmid2000evaluation,Miko2004,Miko2005}, the repeatability
criterion is the {\it de facto} standard procedure to assess keypoint detection
performance~\cite{tuytelaars2008local}.

The repeatability rate measures the detector's ability to identify the same
features (i.e., \emph{repeated} detections) despite variations in the viewing
conditions.
Defined as the ratio between the number of keypoints simultaneously present in
all the images of the series (repeated keypoints) over the total number of
detections, it can be seen as a measure of the detector's efficiency.
Indeed, the repeatability rate incorporates two struggling quality criteria:
the number of repeated detections (i.e., potential correspondences) should be
maximized while the total number of detections should be minimized since the
complexity of the matching grows with the square of the number of detections.

Interest point detectors can also be indirectly evaluated through a
particular application. In~\cite{mikolajczyk2005performance}, the authors
propose to evaluate detector-descriptor combinations in an image
matching/recognition scenario.
Although this approach can lead to very practical observations, the conclusions
about the keypoints stability is intertwined with the descriptor's
discrimination ability.

In this work we show that the repeatability criterion suffers from a systematic
bias: it favors redundant and overlapped detections.
This has serious consequences, as evenly distributed and independent detections
are crucial in image matching applications.
The concentration of many keypoints in a few image regions is generally not
helpful, no matter how robust and repeatable they may be.
%
A performance metric should therefore prioritize detectors giving evenly
distributed keypoints over those giving redundant ones.
To better measure the detectors redundancy, we introduce a modified
repeatability criterion.
We consider the area actually covered by the descriptor
and we evaluate the {\it descriptor overlap} as a measure of redundancy. \\

\noindent{\bf Contributions and plan of the paper:}
Section \ref{sec:repeat:bias} describes the repeatability criterion, discusses its variants, and illustrates how
algorithms with redundant detections and unbalanced spatial distribution may
perform better according to this traditional quality measure.
Section \ref{sec:repeat:coverage} is the core section as it introduces a simple correction of the
repeatability criterion that involves descriptor overlap.  Section
\ref{sec:state:art} reviews twelve state-of-the-art detectors and specifies the elliptical region associated with each 
detection, as it is given in the original papers. This domain will be used for a fair overlap measure. To gain some 
intuition of the problem, the detection maps of the twelve detectors are also displayed on some benchmark images 
and their visual overlap commented.  Comparative performance tables and maps gathered in Section
\ref{sec:experiments} show that the hierarchy of
detectors is drastically altered by the new repeatability criterion. This result is confirmed by a sanity check on the 
detection/matching performance of these detectors where for a fair comparison we use the same descriptor 
technique (SIFT) for all detectors.
Section 6 contains a final discussion.

\section{The repeatability criterion and its bias}\label{sec:repeat:bias}

\subsection{Definition of the repeatability criterion}\label{sub:sec:repeat:formal}
Consider a pair of images $u_a(\bx)$, $u_b(\bx)$ defined for $\bx \in \Omega
\subset \mathbb{R}^2 $ and related by a planar homography $H$, that is,
$u_b=u_a \circ H$.  The detector repeatability rate for the pair
$(u_a,u_b)$ is defined as the ratio between the number of detections
simultaneously present in both images, i.e., repeated detections, and the total
number of detections in the region covered by both images.

In the repeatability framework, a detection generally consists of an elliptical
region, denoted $R(\bx,\Sigma)$, parametrized by its center $\bx$ and a $2\times2$
positive-definite matrix $\Sigma$,
$$
R(\bx,\Sigma) = \left\{ \bx' \in \Omega \mid (\bx'-\bx ) ^T \Sigma^{-1} (\bx'- \bx) \le
1\right\}.
$$
A pair of detections (elliptical regions $R(\bx_a, \Sigma_a)$ and $R(\bx_b,\Sigma_b)$)
from images $u_a(\bx)$ and $u_b(\bx)$ will be considered repeated
if
\begin{equation}
    1 - \frac{ \left\vert  R(\bx_a,\Sigma_a) \cap R(\bx_{ba},  \Sigma_{ba} )   \right\vert }      {\left\vert
    R(\bx_a,\Sigma_a) \cup R( \bx_{ba}, \Sigma_{ba})    \right\vert        } \leq \
    \epsilon_\text{overlap},
    \label{eq:def:repeat}
\end{equation}
where $\bx_{ba} = H \bx_a$, $\Sigma_{ba} = A^{-1} \Sigma_b (A^T)^{-1}$ represents the reprojection of the ellipse on image $u_b$
on the image $u_a$ and $A$ is the local affine approximation of the homography $H$.

The union and intersection of the detected regions are examined on the
reference image $u_a(\bx)$ by projecting the detection on the image $u_b$ into
the image $u_a$.
The union covers an area denoted by $\left\vert R(\bx_a, \Sigma_a) \cup R( \bx_{ba},  \Sigma_{ba})\right\vert$
while  $ \left\vert R(\bx_a, \Sigma_a) \cap R(\bx_{ba},  \Sigma_{ba}) \right\vert$
denotes the area of their intersection.  The parameter
$\epsilon_\text{overlap}$ is the maximum overlap error tolerated.
In most published benchmarks
it is set to $0.40$~\cite{Miko2004,Miko2005,sifer}.

\begin{figure}[hptb]
    \begin{center}
        \includegraphics[width=0.3\textwidth]{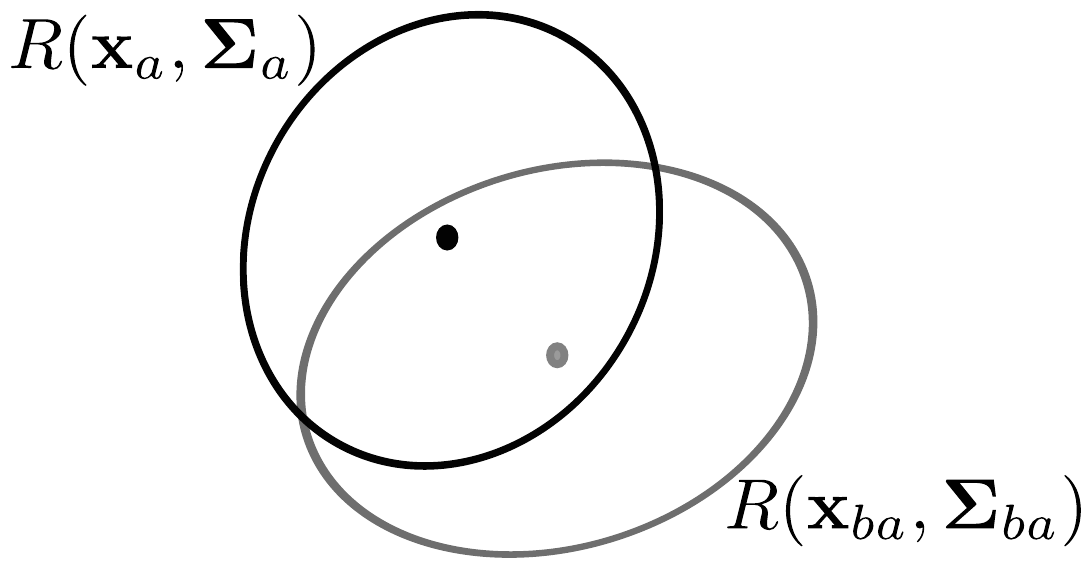} 
    \end{center}
    \caption{
        Illustration of the repeatability criterion.
        Detection $R(\bx_b,\Sigma_b)$ on image $u_b$ is reprojected on the reference image $u_a$.
        If the overlap error is lower than $\epsilon_\text{overlap}$, the detections are considered repeated.
    }\label{fig:illustr:repeat}
\end{figure}

Since the number of repeated detections is upper bounded by the minimal number
of detections, the repeatability rate is defined as
\begin{equation}
    \text{rep} = \frac{ \text{ number of repeated detections}  }{
\min\left( |\mathcal{K}_a|_\Omega ,|\mathcal{K}_b |_\Omega \right) }
\label{eq:rep}
\end{equation}
where  $|\mathcal{K}_a|_\Omega$ and $|\mathcal{K}_b|_\Omega$ denote the
respective numbers of detections inside the area of $\Omega$ covered by both
images $u_a$ and $u_b$.

\subsection{Illustration and alternative definitions}

\begin{figure}[hptb]
    \begin{center}
        \begin{minipage}[c]{0.49\columnwidth}
            \centering
            \includegraphics[width=.85\textwidth]{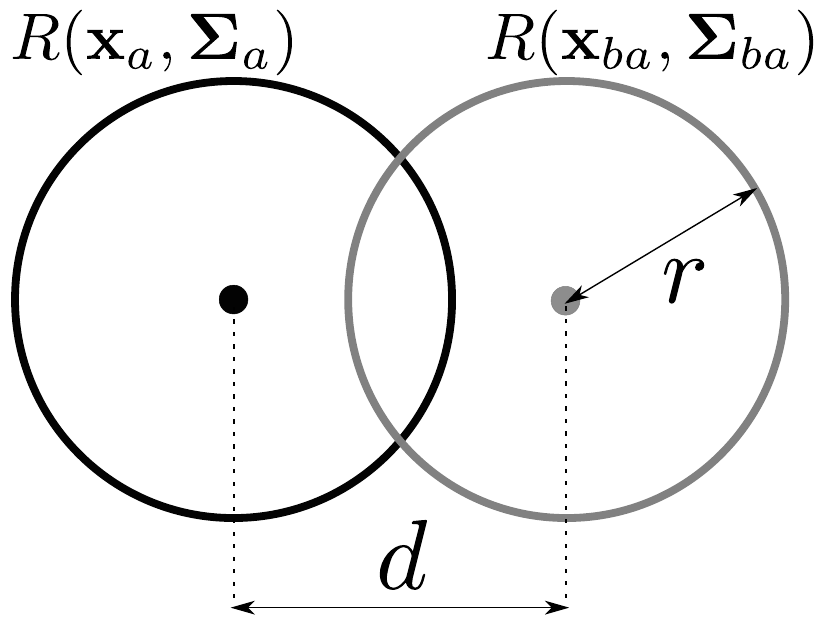}\\
            \vspace{1em}
            \footnotesize {\bf (a)}
        \end{minipage}
        \begin{minipage}[c]{0.49\columnwidth}
            \centering
            \includegraphics[width=.85\textwidth]{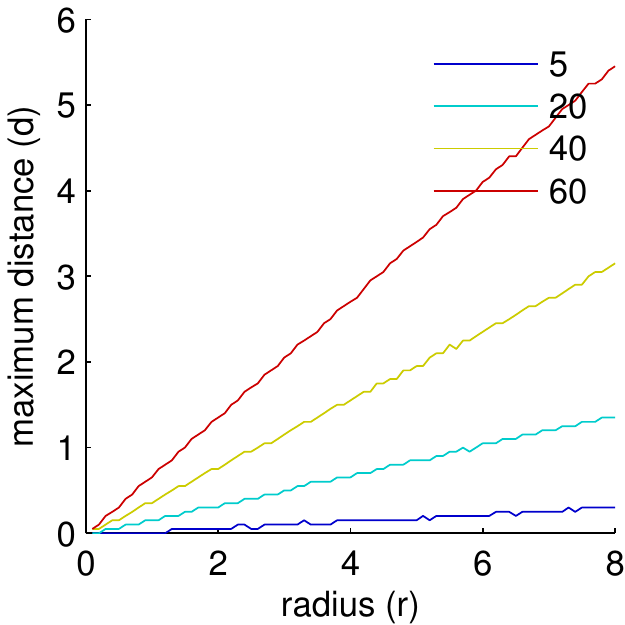}\\
            \footnotesize {\bf (b)}
        \end{minipage}
        \begin{minipage}[c]{0.49\columnwidth}
            \centering
            \includegraphics[width=.85\textwidth]{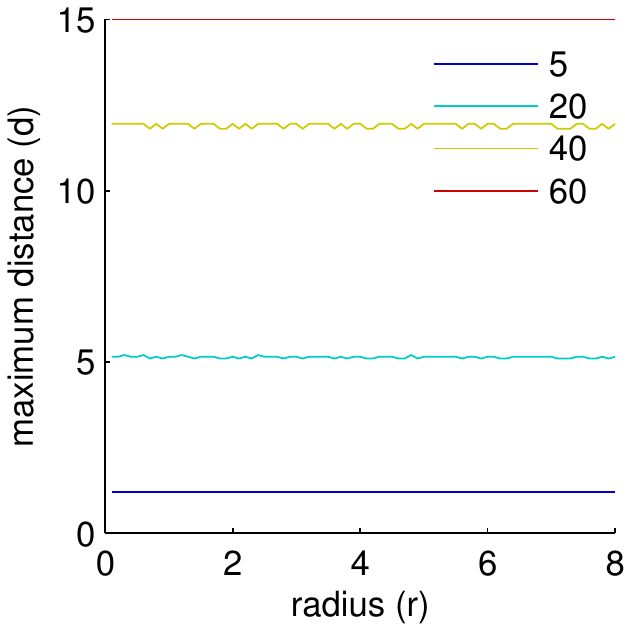}\\
            \footnotesize {\bf (c)}
        \end{minipage}
        \begin{minipage}[c]{0.49\columnwidth}
            \centering
            \includegraphics[width=.85\textwidth]{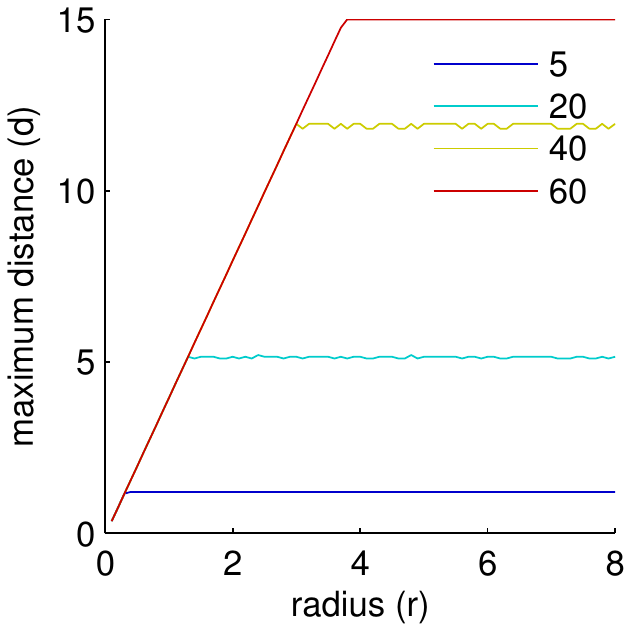}\\
            \footnotesize {\bf (d)}
        \end{minipage}
        \vspace{.3em}
    \end{center}
\caption{
    Illustrating three different definitions of the repeatability criteria.
    Consider a pair of detections whose
    re-projections on the reference image are two disks of radius $r$ with
    their centers separated by $d$ {\bf (a)}. The maximal tolerated distance
    $d_\text{max}$ between repeated detections is plotted as a function of the
    radius $r$ for four values of the parameter $\epsilon_\text{overlap}$
    ($5\%$, $20\%$, $40\%$ and $60\%$).
    {\bf (b)} original definition given by \eqref{eq:def:repeat}, {\bf (c)}
    with ellipses normalization $\kappa=30$, {\bf (d)} definition implemented
    in the provided code provided by the authors of~\cite{Miko2005}. Only the first definition is scale invariant.
}
        \label{fig:repeat:variants}
\end{figure}

To illustrate and discuss the repeatability criterion, let us consider the
particular case of a pair of detections $R(\bx_a,\Sigma_a)$ and $R(\bx_b,\Sigma_b)$
whose re-projections on the reference image are two disks, both of radius $r$
and with centers separated by a distance $d$ (Figure~\ref{fig:repeat:variants}~{\bf
(a)}).
Such a pair will be considered repeated if $d/r \le
f(\epsilon_\text{overlap})$, where $f$ is a monotone function easily derived
from \eqref{eq:def:repeat}.
Figure~\ref{fig:repeat:variants}~{\bf (b)} shows
the maximum distance $d$ under which both detections will be considered repeated
as a function of the radius $r$.

As pointed out in \cite{Miko2005},
detectors providing larger regions have  a better chance of
yielding good overlap scores, boosting as a result their repeatability scores.
This also means that one can artificially increase the repeatability score of any detector by
increasing the scale associated with its detections.

The authors of \cite{Miko2005} proposed to avoid this objection
by normalizing the detected region size before computing the overlapped error.
%
The two detected elliptical regions $R(\bx_a,\Sigma_a)$ and $R(\bx_b,\Sigma_b)$
in~\eqref{eq:def:repeat} are replaced respectively by the elliptical regions $R(\bx_a, \nicefrac{\kappa^2}{r_a R_a} \Sigma_a)$
and $R(\bx_b, \nicefrac{\kappa^2}{r_b R_b} \Sigma_b)$,
 where $r_a$ and $R_a$ are the
radii of the elliptical region $R(\bx_a, \Sigma_a)$ and $\kappa =  30$ is its radii
geometric mean after normalization. 

This normalization prevents boosting a detector's performance by enlarging its associated ellipse.
Yet, such a criterion is not scale-invariant, meaning that it may be over or under
permissive depending on the detection size.
For example, the maximal distance separating repeated detections of equal size
does not take into account the scale (e.g., the radius of the circle in our
special case illustration, see Figure~\ref{fig:repeat:variants} {\bf (c)}). 
In consequence, with $\epsilon_\text{overlap}$ set to its standard value
($\epsilon_\text{overlap} = 40\%$), two circular detections of radius $1$px and
centers separated by $12$px can still be regarded as repeated, although their
respective descriptors may not even overlap!

Surprisingly, the code provided by the authors of
\cite{Miko2005}\footnote{Matlab code
\url{http://www.robots.ox.ac.uk/~vgg/research/affine/} retrieved date} does not
implement the definition presented in their article.
The code introduces a third definition by incorporating an additional
criterion on the maximum distance separating two repeated keypoints that depends on the scale by
$$
| \bx_a - H \bx_b | \leq 4 \sqrt{r_a R_a}.
$$

This criterion is illustrated in Figure~\ref{fig:repeat:variants} {\bf (d)} for
the same study case of two circular detections of equal size. This third criterion is
not scale invariant either.

This explains why we choose to return to the initial scale-invariant definition of
repeatability as given by~ \eqref{eq:def:repeat}.
With the non-redundant repeatability criterion to be introduced in the next section,
it will become pointless  to try ``boosting'' a detector's scale. 
Indeed such attempts will result in decreased matching
performance. The detection's characterizing scale will be
the spatial extent of the descriptor ultimately computed, which is 
the real practical scale associated with each detector. 

\subsection{Repeatability favors redundant detectors}

The following mental experiment illustrates how the repeatability
favors redundancy.
Let DET be a generic keypoint detector, and let DET2 be a variant in which each
detection is computed twice.  The number of repeatable keypoints and the total
number of detections are both artificially doubled, leaving the repeatability
rate unchanged.  However, although the number of costly descriptor computations
has doubled, no extra benefit can be extracted from the enlarged set of
repeated keypoints. The classic repeatability rate fails to report that the benefit
over cost ratio of DET2 is half the one of DET.

This explains why methods producing correlated detections may  misleadingly get
better repeatability ratios. Now the question is: how eliminate the effect of such
correlations?

\section{Non-redundant repeatability}

\label{sec:repeat:coverage}
Besides  the repeatability measure, which ignores the keypoints spatial
distribution, other specific metrics have been proposed. Some examine the spatial 
distribution of the descriptors and others evaluate how well they describe the image.
The ratio between the convex hull of the detected features and the total
image surface is used in~\cite{dickscheid2009evaluating} as a coverage measure.
The harmonic mean of the detections positions is used
in~\cite{ehsan2011measuring,ehsan2013rapid} as a measure of concentration.
In~\cite{dickscheid2011coding}, the authors propose to measure the completeness
of the detected features, namely the ability to preserve the information
contained in an image by the detected features.
The \emph{information content} metric proposed in~\cite{schmid2000evaluation}
quantifies the distinctiveness of a detected feature with respect to the whole
set of detections. Non specific features are indeed  harmful, as they can match to
other many and therefore confuse the matching.
Being complementary to it, these metrics are generally used in combination with the
repeatability rate.
Nevertheless, since the purpose of the repeatability is to report on the
benefit/cost ratio of a given detector, it should also, by itself, report on
the description redundancy.
In fact the descriptors redundancy can be naturally incorporated in the repeatability
criterion.

\subsection{Incorporating descriptor overlap in the repeatability criteria}

To evaluate the redundancy of a set of detections $k\in \mathcal{K}$, each
detection $(\bx_k,\Sigma_k)$ is assigned a mask function $f_k(\bx)$ consisting of a
truncated elliptical Gaussian
$$
    f_k(\bx) = K e^{-\frac{1}{2\zeta^2} (\bx-\bx_k)^T \Sigma^{-1}_k (\bx - \bx)} ,
$$
if $ (\bx-\bx_k)^T \Sigma^{-1}_k (\bx - \bx)  \le \rho^2$ and $0$ elsewhere.
Each mask is normalized so that its integral over the image domain is equal to 1.
The values $\rho$ and $\zeta$ control the extent of the detected feature, as it can be
derived from the descriptor's design. They will be fixed  for  each detector by referring to the original paper where it was introduced (section \ref{sec:state:art}). Indeed most detectors proposals come up with a descriptor or at least with a characterization of the region where this descriptor should be computed. 

%
%

\begin{figure}[hptb]
    \begin{center}
        \begin{minipage}[c]{0.23\columnwidth}
            \centering
            \includegraphics[width=\columnwidth]{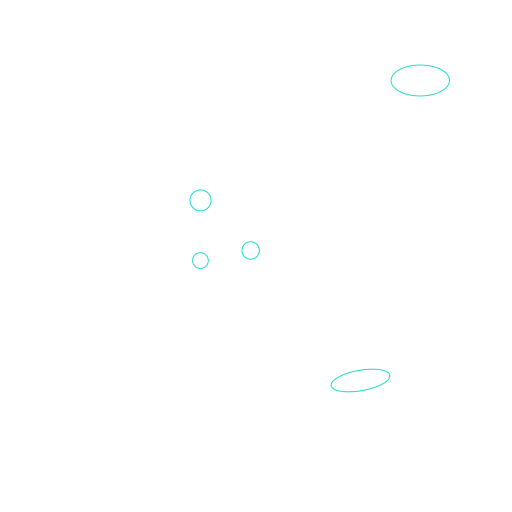}  
            \footnotesize a)
        \end{minipage}
        \begin{minipage}[c]{0.23\columnwidth}
            \centering
            \includegraphics[width=\columnwidth]{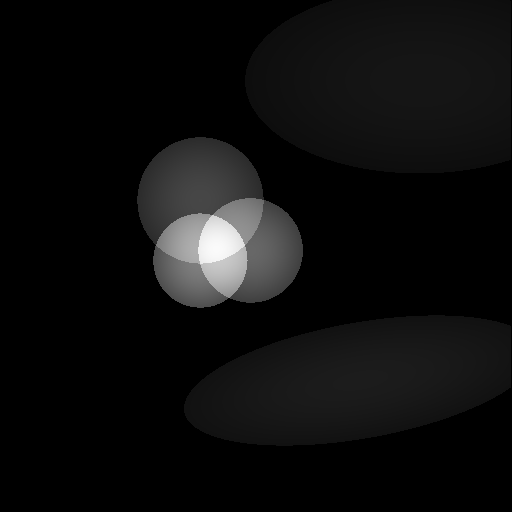}
            \footnotesize b)
        \end{minipage}
        \begin{minipage}[c]{0.23\columnwidth}
            \centering
            \includegraphics[width=\columnwidth]{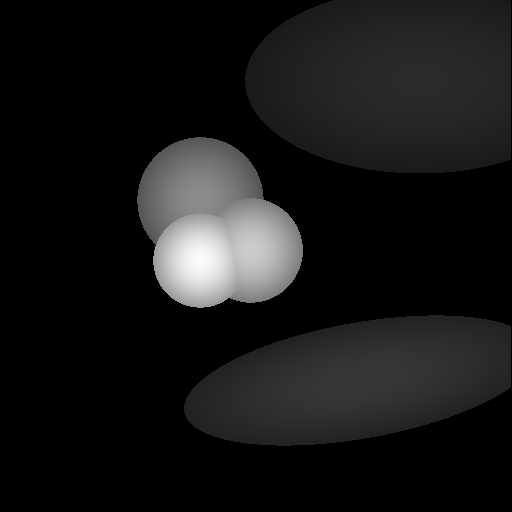}
            \footnotesize c)
        \end{minipage}
        \begin{minipage}[c]{0.23\columnwidth}
            \centering
            \includegraphics[width=\columnwidth]{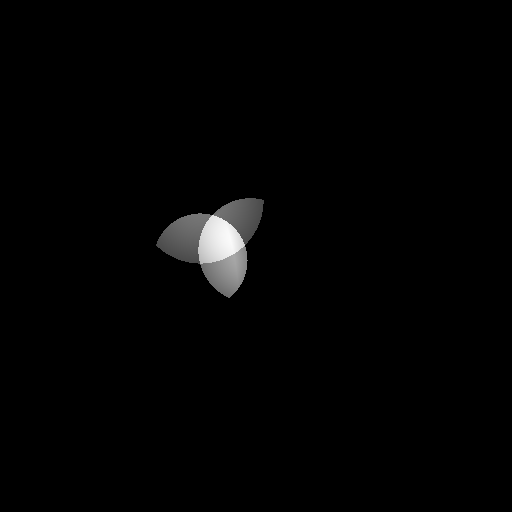}
            \footnotesize d)
        \end{minipage} \vspace{.3em}

\caption{
    The mask functions formalizing the keypoint description on a toy example consisting of several Gaussian blobs (a).
    The sum over all detections $\sum_{k\in \mathcal{K}} f_k(\bx)$ maps the contribution of
    each image pixel to different descriptors (b).
    The max over all detections masks $\max_{k\in \mathcal{K}} f_k(\bx)$ maps the
    pixel contributions to the best available descriptor (c). Their difference  maps the detection redundancy (d).
}
        \label{fig:example:fk}
    \end{center}
\end{figure}

The sum of all descriptor masks $\sum_{k \in \mathcal{K}} f_k(\bx)$ yields a final map
showing how much each image pixel contributes to the set of all computed
descriptors.  Note that  one pixel may contribute to several descriptors (as in
the example shown in Figure~\ref{fig:example:fk}).  Similarly, the maximum
taken over all detections $\max_{k \in \mathcal{K}} f_k(\bx)$ maps the
pixels contribution to the best descriptor. 
Thanks to the mask normalization, the number of keypoints $K := card \left(
\mathcal{K} \right)$ is given by
\begin{equation}
    K = \int_\Omega \left(\sum_{k \in \mathcal{K}} f_k(\bx)\right)d\bx,
\end{equation}
where $\Omega$ denotes the image domain.
On the other hand,
\begin{equation}
K_\text{nr} := \int_\Omega \left(\max_{k \in \mathcal{K}} f_k(\bx)\right) d\bx
\end{equation}
measures the number of {\it non-redundant} keypoints. This value can be
interpreted as a count of the independent detections. To gain some intuition
and see why this measurement is quite natural, let us examine four illustrative
cases. Assume that there are only two detected keypoints so that $K=2$. If the
two detections
\begin{itemize}
\itemsep -.2em

\item completely overlap, then  $K_\text{nr} =  1.$

\item If they share the same center but have different sizes,
    then $1 < K_\text{nr} < K=2$.
    But if their sizes are significantly different, then $K_\text{nr} \approx 2$, which makes sense. Indeed, one of
    them describes a fine detail and the other one a detail at a larger scale.
    Thus, their information contents are roughly independent.

\item If both keypoints are very close to each other then again $1 <
    K_\text{nr} < K=2$ and the above remark on scales still applies.

\item If the descriptors do not overlap at all  then $K_\text{nr}=K = 2$.

\end{itemize}
The propensity of a given algorithm to extract overlapped and redundant
detections can therefore be measured by computing the {\it non-redundant detection
ratio}:
\begin{equation}
    \text{nr-ratio} := K_\text{nr}/K.
    \label{nr:ratio}
\end{equation}

\noindent \textbf{Non-redundant Repeatability.}
The repeatability criterion \eqref{eq:rep} can now be modified to take into account  detection redundancy.
Let $\mathcal{K}_r$ be the set of repeatable keypoints between two snapshots, and  $\Omega$ the area simultaneously covered by both images.
We define the  {\it non-redundant repeatability rate} by
\begin{equation}
    \text{nr-rep} := \frac{  \int_\Omega \max_{k\in \mathcal{K}_r}  f_{k}(\bx) dx dy} {\min\left( |\mathcal{K}_a|_\Omega ,|\mathcal{K}_b |_\Omega \right) }
    \label{eq:nr:rep}
\end{equation}
where  $|\mathcal{K}_a|_\Omega$ and $|\mathcal{K}_b|_\Omega$ denote the respective numbers of detections inside $\Omega$.
The number of repeated detections in
\eqref{eq:rep} is replaced in \eqref{eq:nr:rep} by the number of
non-redundant detections.

\section{The domain of state-of-the-art feature detectors}
\label{sec:state:art}


In this section we review the twelve state-of-the-art feature
detectors that will be compared using the non-redundant repeatability criteria. Our goal is to specify the region of the descriptor associated with each detector. It is classically  objected that the descriptors associated with a detector may influence its matching performance. Hence the descriptor performance should be evaluated independently of its associated descriptor, and conversely. Fortunately, most papers introducing a detector also specify the area of interest around each detector as a circular or elliptical region. This is the region on which the final descriptor will be computed, regardless of its description technique.  This information about the descriptor's region can be taken from the original papers. It is independent of the ultimate choice of a description technique, which may indeed vary strongly. In our discussion of each detector, we shall nevertheless also associate a fixed type of SIFT descriptor to each method, so as to be able to compare matching performance on an equal footing for each method. (This comparison is performed at the end of the experimental section.) A SIFT descriptor can be associated to each elliptical region in a canonical way.

Some of the detectors considered here were also compared in the original
benchmark by Mikolajczyk et al.~\cite{Miko2005}, namely, the Harris-Laplace and
Hessian-Laplace~\cite{Miko2005}, Harris-Affine and
Hessian-Affine~\cite{Miko2005}, EBR~\cite{tuytelaars1999content},
IBR~\cite{tuytelaars2000wide} and MSER~\cite{matas2004robust}.  We also
included here for completeness methods published since:
SIFT~\cite{Lowe1999,Lowe2004}, SURF~\cite{bay2006surf},
SFOP~\cite{forstner2009detecting}, BRISK~\cite{leutenegger2011brisk} and
SIFER~\cite{sifer}.
Table \ref{tab:detectors} summarizes the algorithms invariance properties.
For details, we refer the reader to the original methods
publications and to the survey by Tuytelaars and Mikolajczyk~\cite{tuytelaars2008local}.

Furthermore, we shall show detection maps on pattern images as well as on
several natural photographs to illustrate the behavior of each algorithm.

Most keypoint detection methods share the use of the {\it  Gaussian scale-space}
$u(\bx, \sigma)$
defined by
$$u(\bx,\sigma):= (G_\sigma\ast u)(\bx), \,\,\,
\text{with}\,\,\,
G_\sigma(\bx)=\frac 1{2\pi\sigma}e^{-\frac{\| \bx \|^2}{2\sigma^2}},$$
where
$\sigma$
and
$\bx$
are respectively called the scale and space variables.\\

\begin{table*}[hptb]
\footnotesize
\centering
 \begin{tabular}{lcc|cccc}
    \toprule
                & detects     & feature                & rotation & zoom & homothety & affine  \\\midrule
SIFT            & $(\bx,\sigma)$ & blob                   & yes      & yes  & no        & no      \\
EBR             & parallelograms & corners                & yes      & no   & yes       & limited \\
IBR             & $(\bx,\Sigma)$    & blob                   & yes      & no   & yes       & yes     \\
Hessian-Laplace & $(\bx,\sigma)$ & blob                   & yes      & yes  & no        & no      \\
Hessian-Affine  & $(\bx,\Sigma)$    & blob                   & yes      & yes  & no        & limited \\
Harris-Laplace  & $(\bx,\sigma)$ & corner                 & yes      & yes  & no        & no      \\
Harris-Affine   & $(\bx,\Sigma)$    & corner                 & yes      & yes  & no        & limited \\
MSER            & regions        & contrasted level lines & yes      & no   & no        & yes     \\
SURF            & $(\bx,\sigma)$ & blob                   & limited  & yes  & no        & no      \\
SFOP            & $(\bx,\sigma)$ & junction, circles      & yes      & no   & yes       & no      \\
BRISK           & $(\bx,\sigma)$ & corners                & yes      & yes  & no        & no      \\
SIFER           & $(\bx,\sigma)$ & blob                   & no       & no   & yes       & limited \\
    \midrule
 \end{tabular}
 \vspace{0.5em}
 \caption{}
 \label{tab:detectors}

\vspace{-2em}

\flushleft
 \normalfont\sffamily\normalsize
 Summary of algorithms' invariance properties.
 A zoom is the combination of a homothety and a Gaussian smoothing
 modeling the camera's point spread function.
 The considered detectors detect elliptical regions $(\bx,\Sigma)$,
 circular regions $(\bx,\sigma)$, regions or parallelograms.

\end{table*}

\noindent\textbf{SIFT (scale invariant feature transform)~\cite{Lowe1999,Lowe2004}}
is probably the most popular local image
comparison method. SIFT computes a multi-scale image representation, detects
keypoints from this scale-space, and extracts patch descriptors for each of the
detections.
For detecting keypoints,  SIFT  takes extrema of the convolution of the image
with the normalized Laplacian of Gaussians (LoG).
More precisely, SIFT approximates the LoG kernel by a difference of Gaussians
(DoG),
$$
w_\text{SIFT}(\sigma, \bx) =  \sigma^2 \Delta G_\sigma \ast u(\bx) \approx
\left( G_{k\sigma} - G_\sigma \right) \ast u(\bx),
$$
where $k=2^{\frac1{3}}$ is a constant factor.
The stable interpolated 3D extrema of the multi-scale representation are the
SIFT keypoints.
The description of a keypoint consists of a feature vector assembled from the
gradient distribution over an oriented patch surrounding the detected keypoint.
For a detection at scale $\sigma$, the described patch covers a circular area
of radius $\rho\sigma = 6 \sqrt2 \sigma$ weighted by a Gaussian mask of standard
deviation $\zeta\sigma=6\sigma$ \footnote{In the original SIFT algorithm the
area covered by the descriptor is a  square patch of size $12\sigma \times
12\sigma$. However, to uniformize all the algorithms since some of them do not
give a reference keypoint orientation, we opted to replace the patch by the
smallest disk containing it, which therefore covers a slightly larger area.}.
The described patch is oriented along a dominant orientation of the gradient
distribution.  SIFT considers multiple dominant orientations. This means that
one keypoint may be described by various feature vectors, each corresponding to
one of the dominant orientations.
We shall also consider a variant of SIFT that only takes one feature
vector per detection, the one corresponding to the dominant orientation. We
shall call it SIFT-single (SIFT-S).\\

\noindent\textbf{EBR (edge based regions)~\cite{tuytelaars1999content}} is an
affine-invariant region detector.
This method is not based on a scale-space image representation but on
explicitly searching the image for structures of various sizes.
Starting from a Harris corner point, EBR localizes the two nearby edges and
analyzes their curvature to assign to each segment a characteristic direction
and length.
EBR returns the parallelogram bounded by the two edge
segments. The parallelogram regions can be mapped into elliptical shapes having
the same first and second moments.
The EBR descriptor consists of a set of invariant moments computed over the elliptical region. 
For the sake of comparison, we will rely on the matching experiments on an affine normalized SIFT feature vector computed
over the same elliptical region.
Unlike for the SIFT method, the normalized patch is not weighted by a Gaussian
mask.\\

\noindent\textbf{IBR (intensity based regions)~\cite{tuytelaars2000wide}}
 is an affine-invariant method which detects elliptical shapes of various sizes
 centered on specific gray level extrema.  This method is not based on the
 Gaussian scale-space.
By detecting abrupt changes in the intensity profiles along a set of rays
originating from a gray value extremum, IBR extracts contrasted regions of
various sizes and associates them  elliptical shapes.
Similarly to EBR, invariant moments are computed over the detected
region to build the feature vector.
For a sake of homogeneity in our matching comparisons we shall instead use a SIFT descriptor
computed on the affine normalized patch, without applying a Gaussian weighing mask.\\

\noindent\textbf{Harris-Laplace and Hessian-Laplace detectors~\cite{Miko2005}}.
Unlike SIFT, these methods use two multi-scale representations instead of one.
The first one is used to determine the keypoint location and the second one is
used to select its characteristic scale.
In the case of the Hessian-Laplace method, the first multi-scale representation
is the 2D Hessian determinant while the second one is the normalized Laplacian,
both computed on the Gaussian pyramid~\cite{lindeberg1993scale}.
The 2D Hessian determinant extremum gives the
keypoint location $\bx$. Then, the extremum of the scale-space Laplacian
$\Delta u(\bx,\sigma)$ with respect to $\sigma$ gives the keypoint scale.
The detector goes back and forth between both multi-scale representations to
iteratively refine $\bx$ and $\sigma$.
The Harris-Laplace method proceeds almost identically. Only the Harris
operator~\cite{harris1988combined} is used in place of the 2D Hessian to
extract the keypoint location $\bx$.
The Harris-Laplace features are predominantly corners while the Hessian-Laplace
mostly detects blobs.
Unlike the SIFT method, the extrema are not interpolated to subpixel
precision.
Once extracted, each keypoint is locally described, using the SIFT or the GLOH
descriptor~\cite{Miko2005,mikolajczyk2005performance}.
Consequently, for a detection at scale $\sigma$, the described patch covers a
circular area of radius $\rho\sigma=6 \sqrt2 \sigma$ weighted by a Gaussian
mask of standard deviation $\zeta\sigma=6\sigma$.\\

\noindent\textbf{Harris-Affine and Hessian-Affine detectors~\cite{Miko2005}}
are affine extensions of the Harris-Laplace and Hessian-Laplace detectors.
Instead of detecting keypoints, both methods detect elliptical regions.
Compared to the Harris-Laplace and Hessian-Laplace methods, the affine variants
contain an additional step in which the second-moment matrix is used to
estimate an elliptical shape around each keypoint\footnote{
    The elliptical shape is estimated via an iterative procedure.
    Unreliable detections with degenerated second-moment matrices are also
    discarded in the process.}.
These elliptical shapes are used to normalize the local neighborhood by an affine
transformation before its description (using the SIFT or the GLOH descriptor).
The SIFT descriptor is adopted in the present study.
If $\sigma$ denotes the geometric mean of the ellipse radii, then the described
patch covers a circular area in the affine-normalized neighborhood of radius
$\rho\sigma=6 \sqrt2 \sigma$ weighted by a Gaussian mask of standard deviation
$\zeta\sigma=6\sigma$.\\

\noindent\textbf{MSER (Maximally stable extremal regions)
\cite{matas2004robust}} is an affine-invariant method which extracts regions that are
connected components of image upper level sets.
By examining how the area of the image upper-level sets evolves with
respect to an image intensity threshold, MSER measures the region stability.
The MSERs are the regions that
achieve a local maximum of the (non-positive) derivative of the region area
with respect to its level.
MSER proposes to compute feature descriptors at different scales of
the detected region size (1.5, 2 and 3 times the convex hull of the
detected region). In addition, MSER regions can be easily mapped into
elliptical shapes and then used to compute an affine descriptor of the
detected region. In the present framework, for each of the detected
regions a SIFT feature vector on an affine normalized patch of twice the
size of the detected region was computed.\\

\noindent\textbf{SURF (speeded-up robust features)~\cite{bay2006surf}} can be
regarded as a fast alternative to SIFT.  SURF keypoints are the 3D extrema of a
multi-scale image representation that approximates the 2D Hessian determinant
computed on each scale of the Gaussian scale-space. The Gaussian convolution is
approximated using box filters computed via integral images.
SURF descriptors are computed over a Gaussian window centered at the keypoint,
and encode the gradient distribution around the keypoint using 2D Haar
wavelets.
The described patch for a detection at scale $\sigma$ covers a circular area of
radius $\rho\sigma=10 \sqrt2 \sigma$ weighted by a Gaussian mask of standard
deviation $\zeta\sigma=3.3\sigma$.
Note that the described areas used in SIFT and SURF are slightly different. A SURF descriptor
patch is larger but uses a more concentrated Gaussian mask.\\

\noindent\textbf{SFOP (scale-invariant feature
operator)~\cite{forstner2009detecting}}. SFOP is a versatile multi-scale
keypoint detector that explicitly models and detects corners, junctions and
circular features. SFOP is built on the F\"orstner feature
operator~\cite{forstner1994framework} for detecting junctions and on the spiral
model~\cite{bigun1990structure} for unifying different feature types into a
common mathematical formulation.
For detecting keypoints at different scales, the input image is decomposed into
series of images using a Gaussian pyramid. Each image is then scanned for
various feature types, namely, circular structures of various sizes and
junctions of different orientations.
At each pixel, the algorithm takes a surrounding patch and evaluates its
consistency to the feature model.
Although SFOP only concerns keypoint detection,  the authors recommend
combining the SFOP detector with SIFT's descriptor.
Consequently, the described patch for a detection at scale $\sigma$ also covers
a circular area of radius $\rho\sigma=6 \sqrt2 \sigma$ weighted by a Gaussian
mask of standard deviation $\zeta\sigma=6\sigma$.\\

\noindent\textbf{BRISK (binary robust invariant scalable
keypoints)~\cite{leutenegger2011brisk}} focuses on speed and efficiency.
The BRISK detector is a multi-scale adaptation of  FAST and its optimized
version AGAST~\cite{rosten2006machine,mair2010adaptive} corner detectors.
The AGAST corner detector is first applied separately to each scale of a Gaussian
pyramid decomposition to rapidly identify  potential regions of interests. For
each pixel in such regions,  a corner score quantifying the detection
confidence  is computed (see \cite{mair2010adaptive} for details).
Based on the AGAST corner score, BRISK performs a 3D non-maxima suppression and
a series of quadratic interpolations to extract the BRISK keypoints $(\bx,s)$,
being $(\bx)$ the 2D position and $s$ the feature size.
The BRISK descriptor is a binary string resulting from brightness differences
computed around the keypoint.

In the current analysis, we calibrated the size of the detections $s$ provided by
the BRISK binary to make it comparable to the other methods.
We empirically found that the image of Gaussian of standard deviation $\sigma$
produces a SIFT detection of scale $\sigma$ while it produces a BRISK feature
of size $s=4\sigma$.
In consequence, for a BRISK detection of size $s$, the
described patch in the present study covers a circular area of radius $\rho s = \frac{3}2\sqrt{2} s$
weighted by a Gaussian mask of standard deviation $\zeta s = \frac{3}2 s$.\\

\begin{figure}[hptb]
    \begin{center}
        \begin{minipage}[c]{0.38\columnwidth}
            \centering
            \includegraphics[width=.85\textwidth]{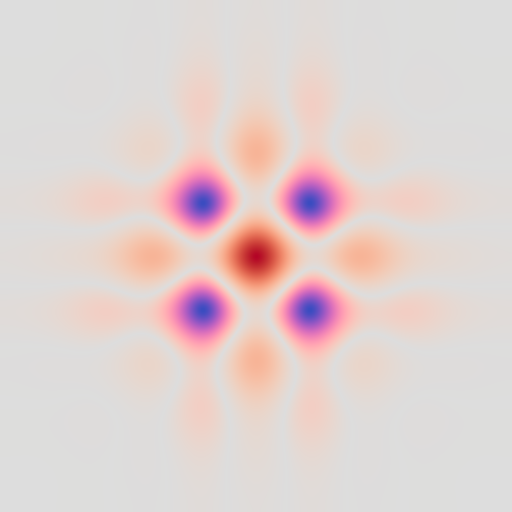}
            \footnotesize $\text{cmg}_\sigma(x,y)$
        \end{minipage}
        \begin{minipage}[c]{0.38\columnwidth}
            \centering
            \includegraphics[width=.85\textwidth]{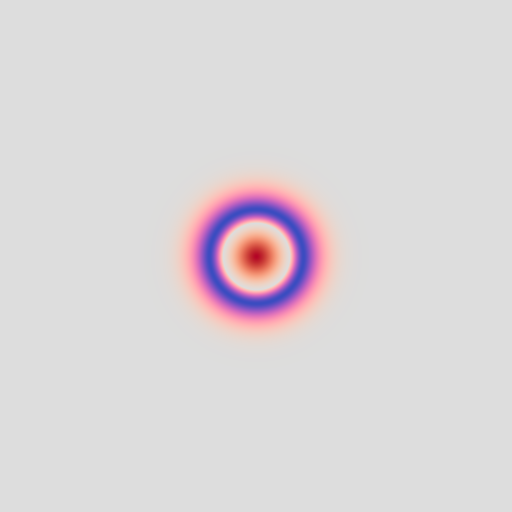}
            \footnotesize $ \sigma^2\Delta G_\sigma(x,y) $
        \end{minipage}
        \vspace{.3em}
    \end{center}
\caption{
    SIFER (left) and SIFT (right) filter kernels.  The SIFER
    kernel, a Gaussian modulated along the two axes by cosine
    functions is  not rotation invariant, while the difference of
    Gaussians used in SIFT is.
}
        \label{fig:kernel:sifer:sift}
\end{figure}

\noindent\textbf{SIFER (scale-invariant feature detector with error
resilience)~\cite{sifer}}.
The recently introduced SIFER algorithm tightly follows SIFT, but computes a
different multi-scale image representation.
Instead of smoothing the image with a set of Gaussian filters and computing its
Laplacian, SIFER convolves the image with a bank of cosine modulated Gaussian
kernels (see Figure \ref{fig:kernel:sifer:sift}).
\begin{equation} \text{cmg}_\sigma(x,y) = \left(2\pi\sigma^2 \left( \cos
\left(\frac{cx}\sigma\right) + \cos\left(\frac{cy}\sigma\right)  \right)
G_\sigma \right).  \end{equation}
The 3D extrema of the resulting multi-scale representation are the SIFER
keypoints.
The method is homothety invariant. Unlike SIFT, however, SIFER is not zoom-out invariant. Indeed, its kernel does not commute
with a Gaussian camera blur.
The authors claim that, despite loosing rotation invariance, the approach
increases the detection precision in both scale and space thanks to the better
localization of the modulated cosine filters.
The descriptor computed at each extracted keypoint is identical to the SIFT
descriptor.
Therefore, the described patch considered in the present study covers a circular
area of radius $\rho\sigma=6 \sqrt2 \sigma$ weighted by a Gaussian mask of
standard deviation $\zeta\sigma=6\sigma$.\\

\subsection{Detection maps}\label{sec:maps}

\begin{figure*}[hptb]
    \centering
    \footnotesize
\begin{minipage}[c]{0.16\textwidth}
\centering
\includegraphics[width=\columnwidth]{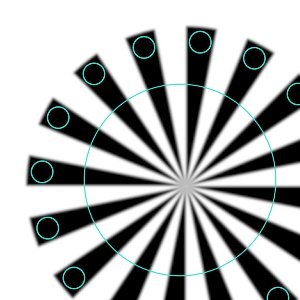}
SIFT (17)\vspace{0.3em}\\
\end{minipage}
\begin{minipage}[c]{0.16\textwidth}
\centering
\includegraphics[width=\columnwidth]{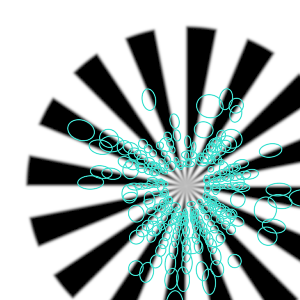}
EBR (249)\vspace{0.3em}
\end{minipage}
\begin{minipage}[c]{0.16\textwidth}
\centering
\includegraphics[width=\columnwidth]{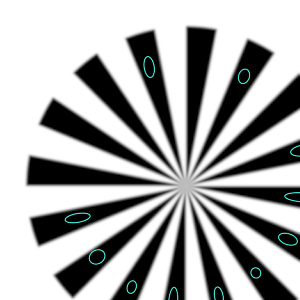}
IBR (13)\vspace{0.3em}\\
\end{minipage}
\begin{minipage}[c]{0.16\textwidth}
\centering
\includegraphics[width=\columnwidth]{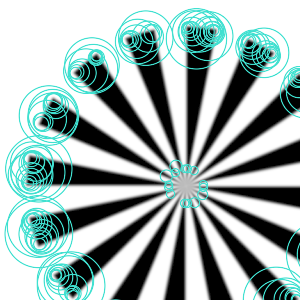}
Harris-Laplace (242)\vspace{0.3em}\\
\end{minipage}
\begin{minipage}[c]{0.16\textwidth}
\centering
\includegraphics[width=\columnwidth]{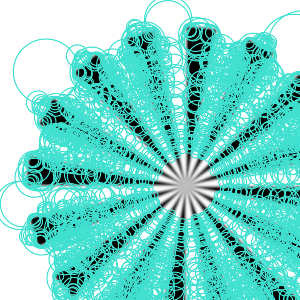}
Hessian-Laplace (1927)\vspace{0.3em}\\
\end{minipage}
\begin{minipage}[c]{0.16\textwidth}
\centering
\includegraphics[width=\columnwidth]{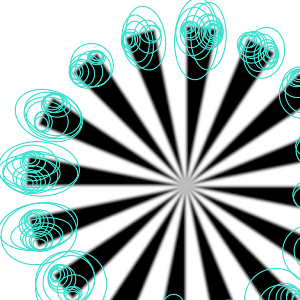}
Harris-Affine (227)\vspace{0.3em}\\
\end{minipage}
\begin{minipage}[c]{0.16\textwidth}
\centering
\includegraphics[width=\columnwidth]{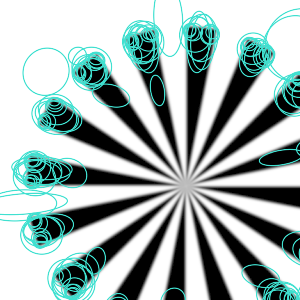}
Hessian-Affine (244)\vspace{0.3em}
\end{minipage}
\begin{minipage}[c]{0.16\textwidth}
\centering
\includegraphics[width=\columnwidth]{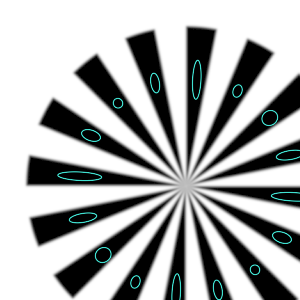}
MSER (18)\vspace{0.3em}
\end{minipage}
\begin{minipage}[c]{0.16\textwidth}
\centering
\includegraphics[width=\columnwidth]{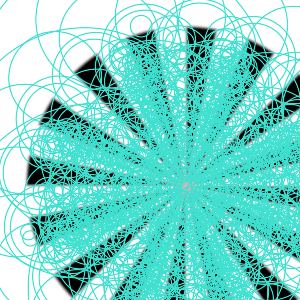}
SURF (652)\vspace{0.3em}\\
\end{minipage}
\begin{minipage}[c]{0.16\textwidth}
\centering
\includegraphics[width=\columnwidth]{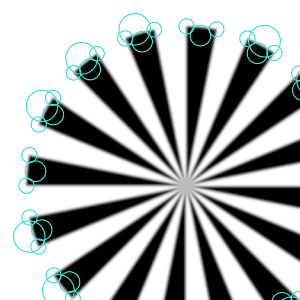}
SFOP (59)\vspace{0.3em}
\end{minipage}
\begin{minipage}[c]{0.16\textwidth}
\centering
\includegraphics[width=\columnwidth]{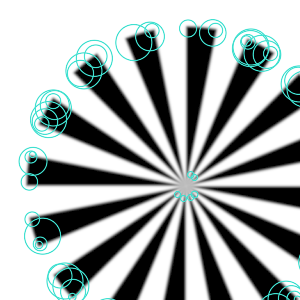}
BRISK (97) \vspace{0.3em}\\
\end{minipage}
\begin{minipage}[c]{0.16\textwidth}
\centering
\includegraphics[width=\columnwidth]{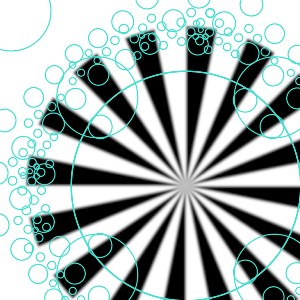}
SIFER (203)\vspace{0.3em}\\
\end{minipage}
\caption{
    Keypoints map on \texttt{siemens star} test image. For  a better readability of the figure, the descriptor
    ellipses are reduced to one sixth of their real size. Thus, when two ellipses overlap, their associated descriptors are in strong overlap. This is particularly conspicuous for the Hessian and Harris  detectors. The total number of
    detected keypoints by each method is shown in brackets.
    SIFT and SFOP seem to be the only (experimentally) rotationally invariant
    methods. The elliptical shapes deduced from the MSER regions have different
    sizes in each rotated triangle.
    By design, SIFT detects blob like structures and SFOP 
    additional features, such as corners and edges.
}
\label{fig:maps:siemens}
\end{figure*}

\begin{figure*}[hptb]
\footnotesize
\centering

\begin{minipage}[c]{0.16\textwidth}
\centering
\includegraphics[width=\columnwidth]{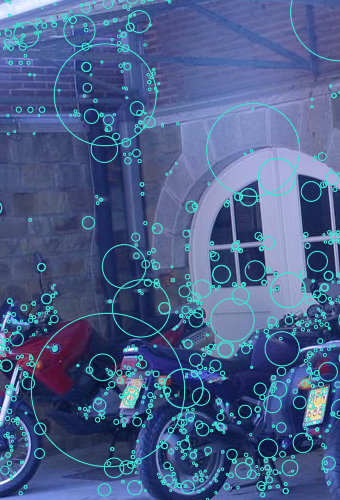}
SIFT (2038) \vspace{0.3em}\\
\end{minipage}
\begin{minipage}[c]{0.16\textwidth}
\centering
\includegraphics[width=\columnwidth]{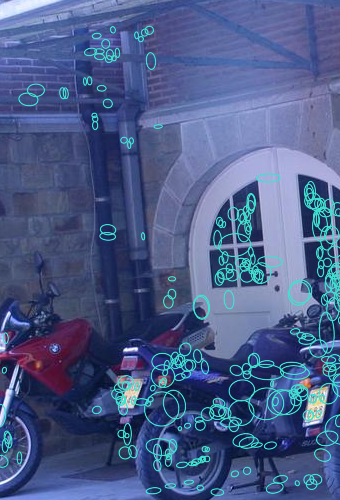}
EBR (644) \vspace{0.3em}
\end{minipage}
\vspace{0.7em}
\begin{minipage}[c]{0.16\textwidth}
\centering
\includegraphics[width=\columnwidth]{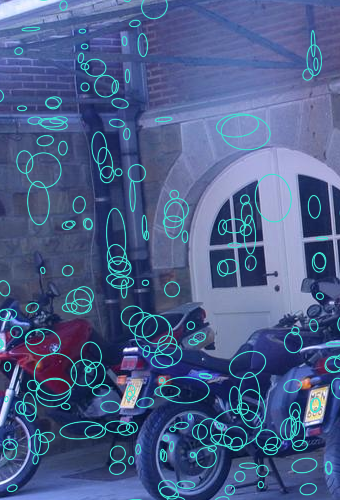}
IBR (652)\vspace{0.3em}\\
\end{minipage}
\begin{minipage}[c]{0.16\textwidth}
\centering
\includegraphics[width=\columnwidth]{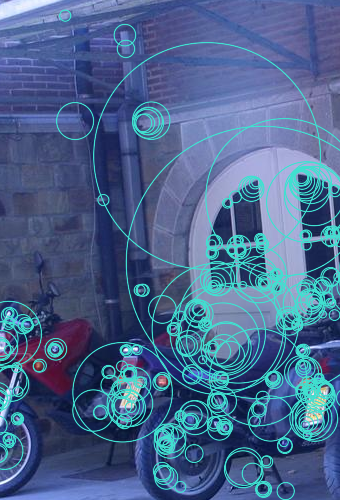}
Harris-Laplace (740)\vspace{0.3em}\\
\end{minipage}
\begin{minipage}[c]{0.16\textwidth}
\centering
\includegraphics[width=\columnwidth]{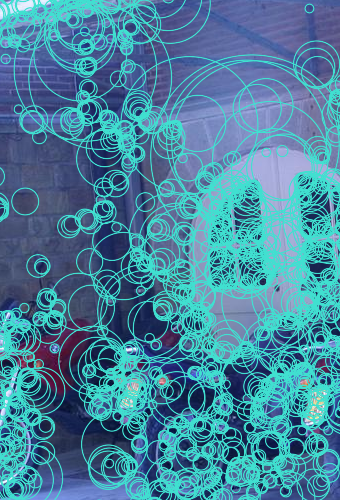}
Hessian-Laplace (3502) \vspace{0.3em}\\
\end{minipage}
\begin{minipage}[c]{0.16\textwidth}
\centering
\includegraphics[width=\columnwidth]{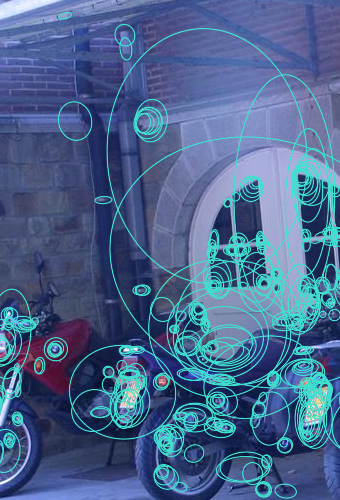}
Harris-Affine (727)\vspace{0.3em}\\
\end{minipage}
\begin{minipage}[c]{0.16\textwidth}
\centering
\includegraphics[width=\columnwidth]{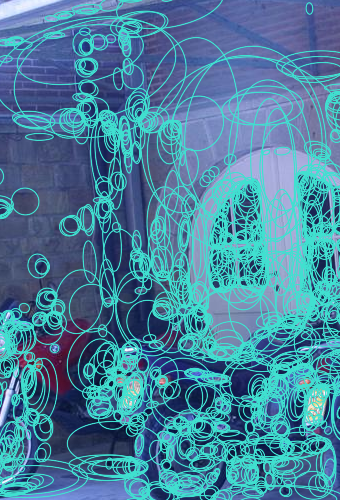}
Hessian-Affine (2857) \vspace{0.3em}\\
\end{minipage}
\begin{minipage}[c]{0.16\textwidth}
\centering
\includegraphics[width=\columnwidth]{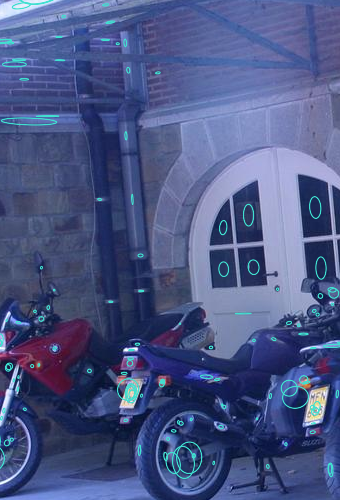}
MSER (352)\vspace{0.3em}\\
\end{minipage}
\begin{minipage}[c]{0.16\textwidth}
\centering
\includegraphics[width=\columnwidth]{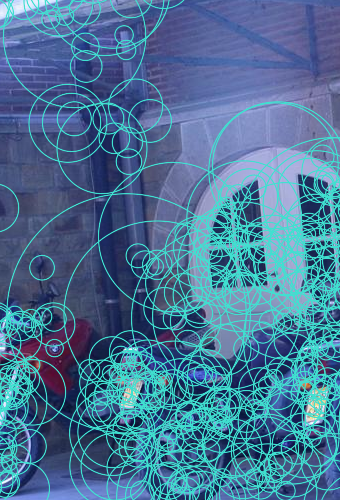}
SURF (781) \vspace{0.3em}\\
\end{minipage}
\begin{minipage}[c]{0.16\textwidth}
\centering
\includegraphics[width=\columnwidth]{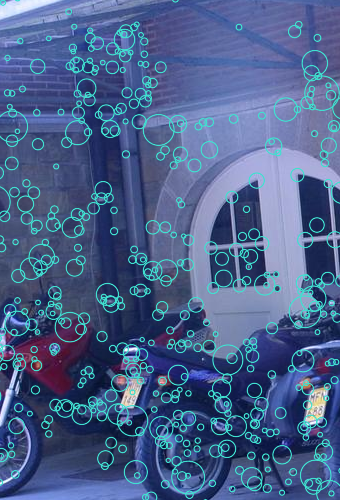}
SFOP (1379)\vspace{0.3em}\\
\end{minipage}
\begin{minipage}[c]{0.16\textwidth}
\centering
\includegraphics[width=\columnwidth]{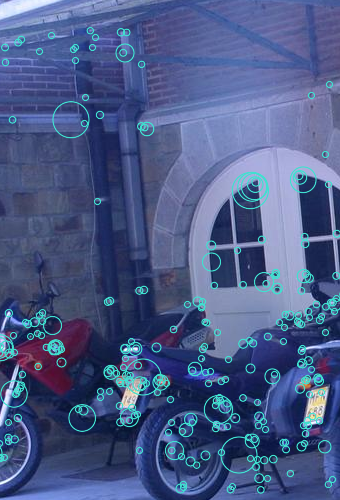}
BRISK (339)\vspace{0.3em}
\end{minipage}
\begin{minipage}[c]{0.16\textwidth}
\centering
\includegraphics[width=\columnwidth]{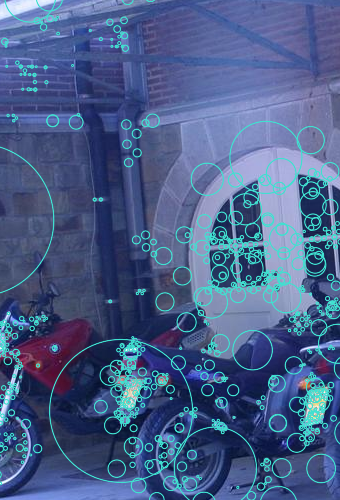}
SIFER (664)\vspace{0.3em}\\
\end{minipage}
\caption{
    Keypoints map on an image from the \texttt{bikes} sequence. For  a better readability of the figure, we reduced six times the descriptors
    ellipses with respect to their real size. This also means that when two ellipses overlap, their associated descriptors are in strong overlap.  The total number of detected keypoints by each method is
    shown in brackets.
    The number of detections significantly varies with the algorithm.  Hessian
    based methods and SIFT produce many more detections than the rest.
    All methods, with the exception of IBR and EBR, detect features at very different scales.
    In particular, SIFT and SFOP detect very small structures.
    Most algorithms detect the same structure several times, producing significantly overlapped detections.
    The SIFER detections are disturbingly concentrated on clusters not necessarily
    overlapped. Yet the proposed non-redundant repeatability metric will not penalize such behavior.
    For the Harris and Hessian based methods, note how corners generate
    trails of detections of increasing size.
}

\label{fig:maps:bikes}
\end{figure*}

Different detectors extract different kind of features, in
different amounts and with different spatial distributions.
To visually inspect the algorithms general behavior, figures
\ref{fig:maps:siemens} to \ref{fig:maps:bikes} show the detection maps for the
twelve compared methods on two pattern images and three images from
the Oxford dataset~\cite{Miko2005} (namely, \texttt{graf}, \texttt{boat} and \texttt{bikes}
sequences).

The detection number varies from one method to the other, and also from one sequence to the next.
MSER generally detects fewer features than the rest while SIFT and the Harris
and Hessian based methods detect many more.

The rotation invariance of the methods is easily  tested by examining the detections
on the \texttt{siemens star} test image shown in  Figure~\ref{fig:maps:siemens}.
Unsurprisingly, SIFT and SFOP are rotation invariant while SIFER is not.
More surprisingly, the Hessian and Harris based methods
are not rotation invariant. Although the Hessian determinant and the Laplacian
of the Gaussian smoothing are isotropic, the methods fail to maintain
the theoretical invariance properties due to the discretization of the differential operators.

Several feature detectors generate multiple detections from a singe local
feature. This is  clearly the case for Harris-Affine, Hessian-Affine and, to a
lesser extent, for BRISK.
In general, with the exception of SIFT, SFOP and MSER, all the detectors appear
to be visually highly redundant.

In some cases, while detections are numerous, they
cluster on a reduced part of the scene.
This is observed for instance with SIFER, (see e.g., Figure~\ref{fig:maps:bikes}).
This seems to imply that the information contained in
the descriptors computed from SIFER keypoints is both redundant and incomplete.


\section{Experiments}
\label{sec:experiments}

\begin{figure*}[hptb]
    \centering
\begin{minipage}[c]{.4975\textwidth}
   \centering

    \includegraphics[width=\textwidth]{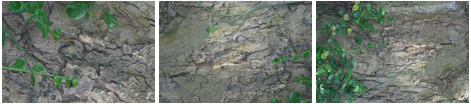}

    \includegraphics[width=\textwidth]{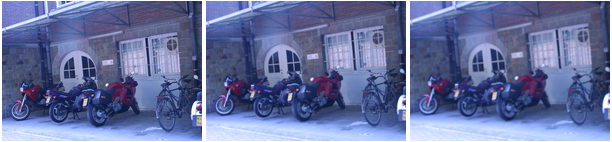}

    \includegraphics[width=\textwidth]{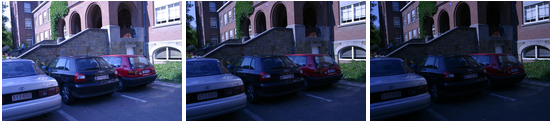}

    \includegraphics[width=\textwidth]{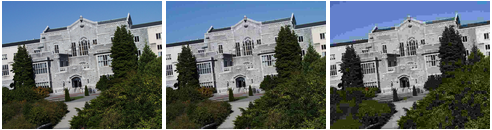}

\end{minipage}
\begin{minipage}[c]{.4625\textwidth}
   \centering
    \includegraphics[width=\textwidth]{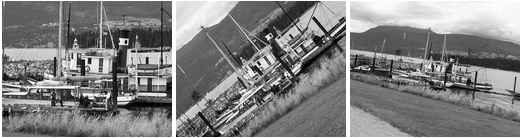}

    \includegraphics[width=\textwidth]{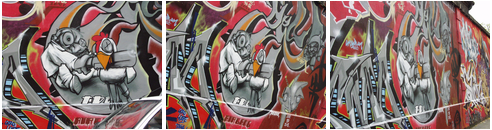}

    \includegraphics[width=\textwidth]{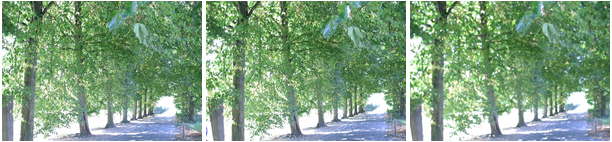}

    \includegraphics[width=\textwidth]{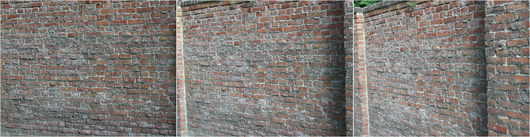}

\end{minipage}
\caption{
     The Oxford dataset.
     From left to right, top to bottom,  \texttt{bark} and \texttt{boat} (scale changes and
     rotations), \texttt{bikes} (camera blur), \texttt{graf} (viewpoint
     changes), \texttt{leuven} (illumination changes), \texttt{trees} camera
     blur, \texttt{ubc} (JPEG compression), \texttt{wall} (viewpoint changes).
}
\label{fig:oxford}
\end{figure*}

To illustrate the proposed non-redundant repeatability criterion, we will examine the
performance of the described feature detectors on the Oxford
dataset~\cite{Miko2005}\footnote{Dataset available at
\url{http://www.robots.ox.ac.uk/~vgg/research/affine/}}.
The Oxford dataset contains eight sequences of six images each designed to help
assess the stability of the detections with respect to habitual image
perturbations, namely, rotation and scale changes, viewpoint changes, camera
blur, illuminations changes and JPEG compression artefacts.
The eight sequences are shown in Figure~\ref{fig:oxford}.
The original and publicly available binaries of all but one methods were
used\footnote{Methods binaries
\url{http://www.robots.ox.ac.uk/~vgg/research/affine/},
\url{http://docs.opencv.org/doc/tutorials/features2d/feature_detection/feature_detection.html},
\url{http://www.vision.ee.ethz.ch/~surf/} and
\url{http://www.cs.ubc.ca/~lowe/keypoints/}
\url{http://www.ipb.uni-bonn.de/sfop/}}.
No reference implementation of SIFER was available, we therefore relied on our
own implementation rigorously following the published description~\cite{sifer}.
The parameters of each method were set to their default values.
All scripts and codes are available for download~\footnote{In particular a documented and optimized
version of the repeatability criteria~\cite{Miko2005} along
with the two variants discussed in Section~\ref{sec:repeat:bias} are available
for download at
\url{http://dev.ipol.im/~reyotero/comparing\_20140906.tar.gz}.}.

The performance evaluation of a detector is two-dimensional.
On the one hand, a detector should produce as many detections as possible,
while on the other, it should keep to a minimum the number of non-repeatable detections.
In other words, the best detector is the one that has simultaneously the largest repeatability
ratio and the largest number of detections.

As we showed in the previous section, a quick visual examination of the
detection maps already reveals that some methods are more redundant
than others. For example, it is clear from Figure~\ref{fig:maps:siemens}
(\texttt{siemens star}) that SIFER, SURF and the Hessian based methods
produce highly redundant detections.
The non-redundancy ratio shown in Table~\ref{tab:all} {\bf (a)} for the eight Oxford sequences helps rank
the methods in terms of redundancy.
With non-redundant ratios lower than $7\%$ on all eight sequences, the
Hessian based detectors are the most redundant methods.
On the other end of the spectrum, the least redundant method is MSER
having an average non-redundant ratio of $51\%$. SIFT and its SIFT-single variant come second, with non-redundant ratios ranging from
$20\%$ to $36\%$.
Since the number of detections of SIFT and Hessian-Laplace are comparable
(Table~\ref{tab:all}~{\bf (b)}), the cost of extracting and matching descriptors
is similar for both methods. Notwithstanding this fact, SIFT produces well-spread
detections while the Hessian-Laplace are redundant and overlapped.
Under such circumstances, we expect that taking into account the descriptors
overlap will change significantly the hierarchy given by the repeatability
rates.

\begin{figure*}[htpb]
    \begin{center}
        \small
        \begin{minipage}[c]{0.49\columnwidth}
            \centering
            \includegraphics[width=.8\textwidth]{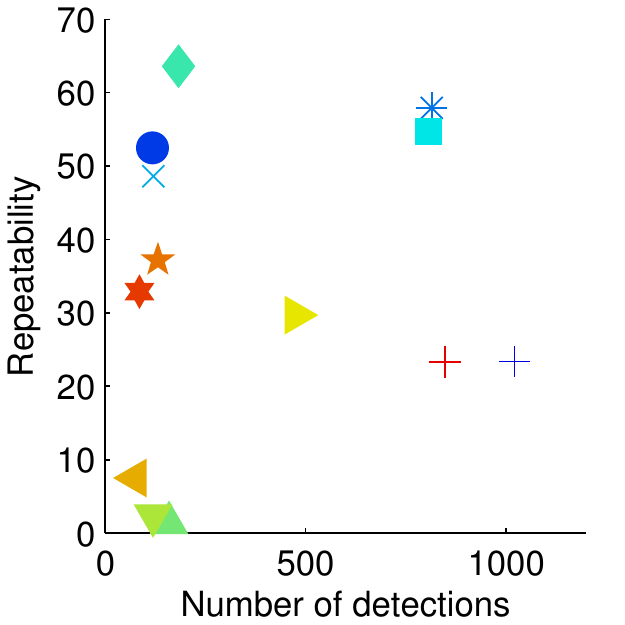}
        \end{minipage}
        \begin{minipage}[c]{0.49\columnwidth}
            \centering
            \includegraphics[width=.8\textwidth]{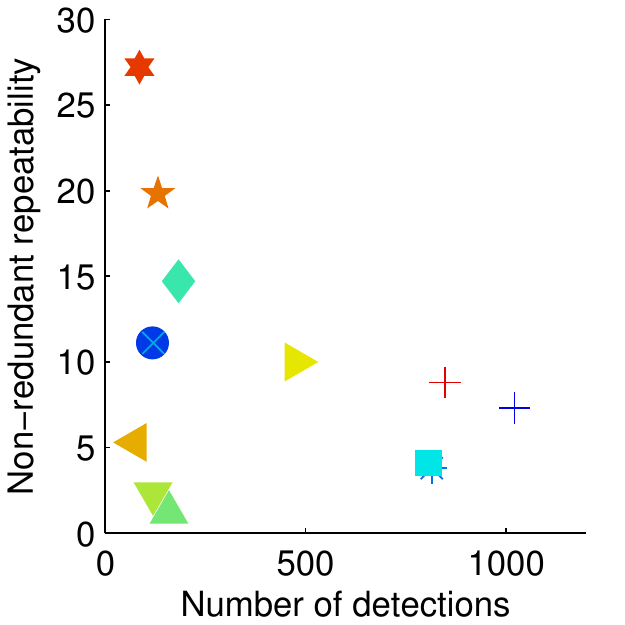}
        \end{minipage}
        \hspace{.6em}
        \begin{minipage}[c]{0.49\columnwidth}
            \centering
            \includegraphics[width=.8\textwidth]{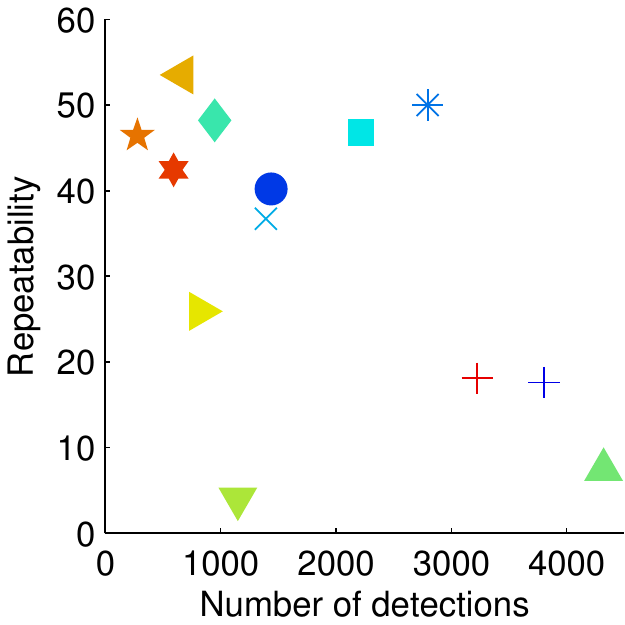}
        \end{minipage}
        \begin{minipage}[c]{0.49\columnwidth}
            \centering
            \includegraphics[width=.8\textwidth]{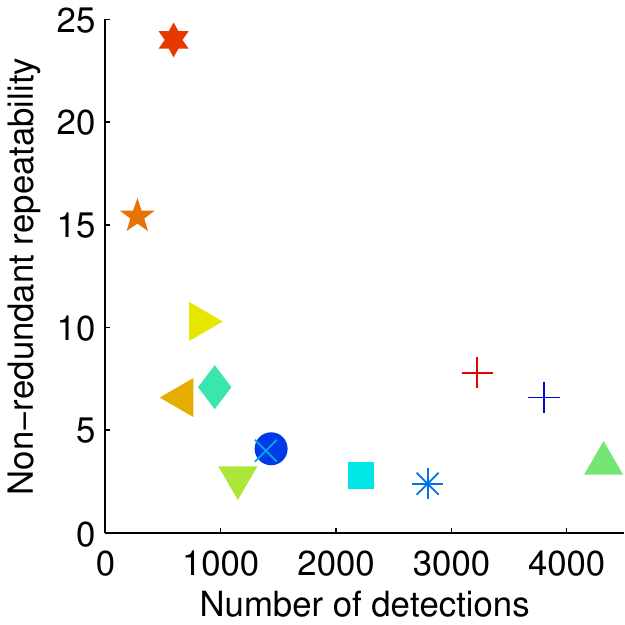}
        \end{minipage}

        \vspace{1.0em}
        \begin{minipage}[c]{0.98\columnwidth}
            \centering
            {\bf (a)} \texttt{bark}  \emph{(scale)}
        \end{minipage}
        \begin{minipage}[c]{0.98\columnwidth}
            \centering
            {\bf (b)} \texttt{boat}  \emph{(scale)}
        \end{minipage}
        \vspace{1.0em}

        \begin{minipage}[c]{0.49\columnwidth}
            \centering
            \includegraphics[width=.8\textwidth]{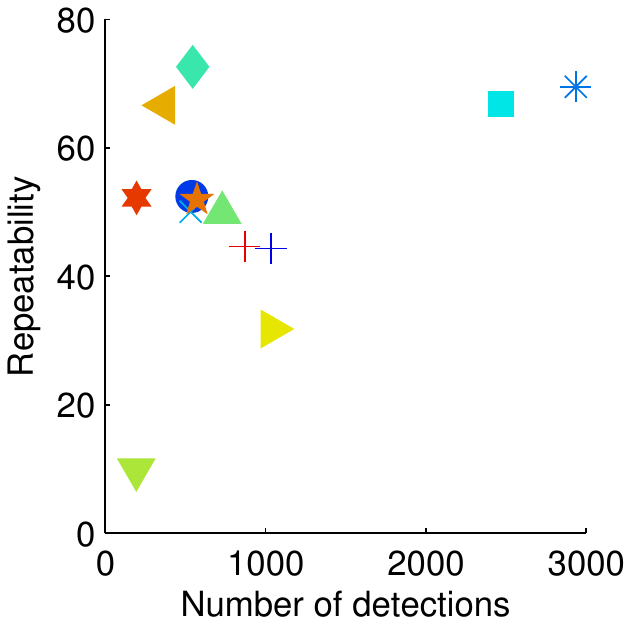}
        \end{minipage}
        \begin{minipage}[c]{0.49\columnwidth}
            \centering
            \includegraphics[width=.8\textwidth]{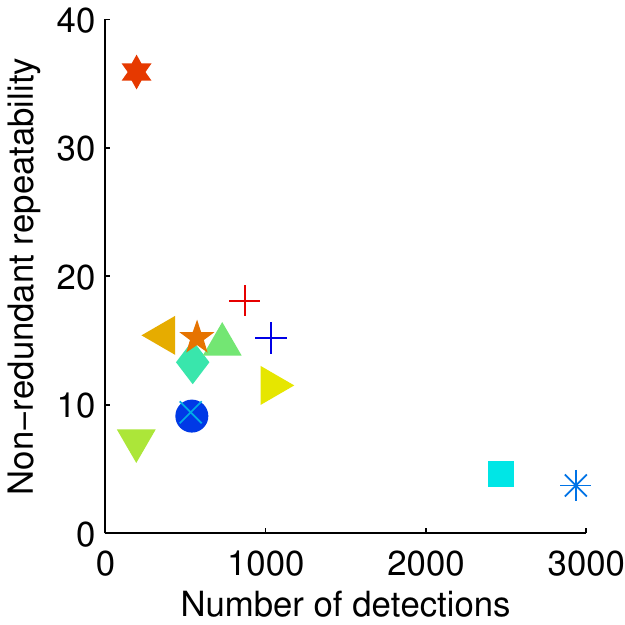}
        \end{minipage}
        \hspace{.6em}
        \begin{minipage}[c]{0.49\columnwidth}
            \centering
            \includegraphics[width=.8\textwidth]{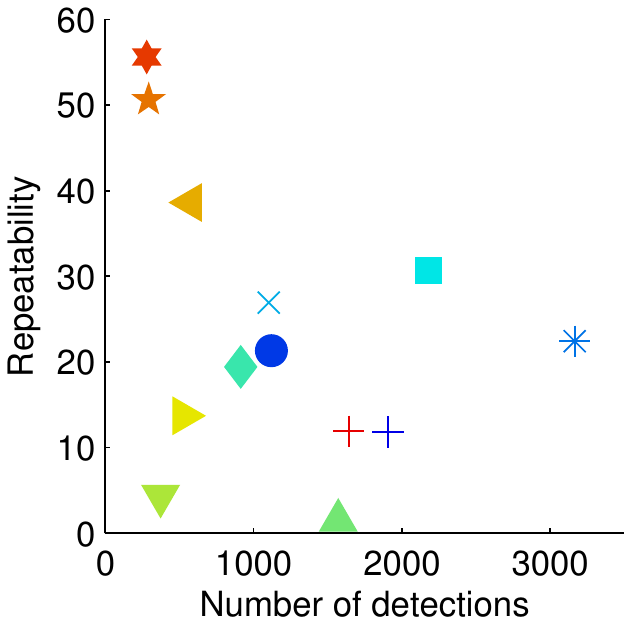}
        \end{minipage}
        \begin{minipage}[c]{0.49\columnwidth}
            \centering
            \includegraphics[width=.8\textwidth]{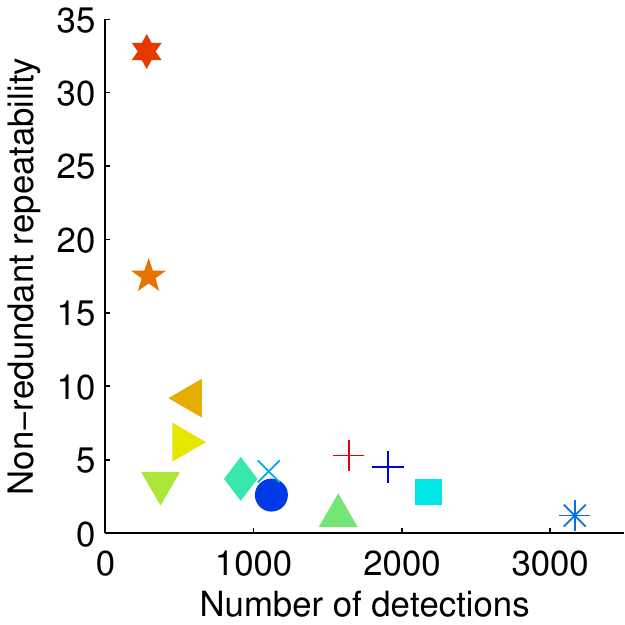}
        \end{minipage}

        \vspace{1.0em}
        \begin{minipage}[c]{0.98\columnwidth}
            \centering
            {\bf (c)} \texttt{bikes}  \emph{(blur)}
        \end{minipage}
        \begin{minipage}[c]{0.98\columnwidth}
            \centering
            {\bf (d)} \texttt{graf}  \emph{(viewpoint)}
        \end{minipage}
        \vspace{1.0em}

        \begin{minipage}[c]{0.49\columnwidth}
            \centering
            \includegraphics[width=.8\textwidth]{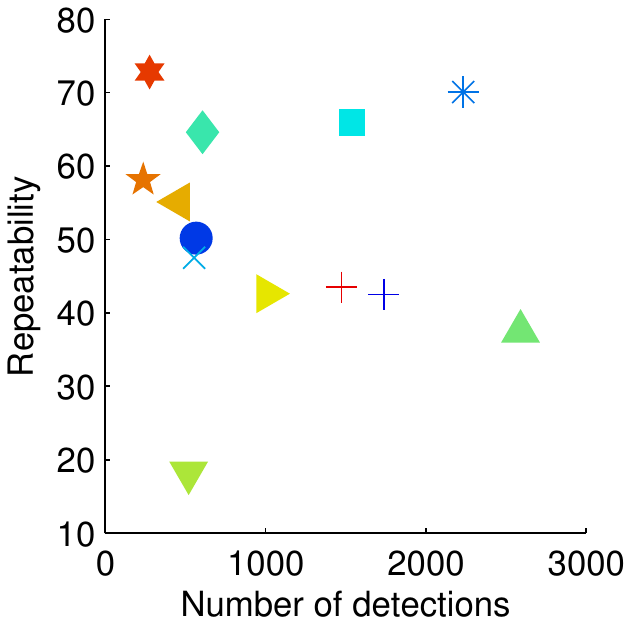}
        \end{minipage}
        \begin{minipage}[c]{0.49\columnwidth}
            \centering
            \includegraphics[width=.8\textwidth]{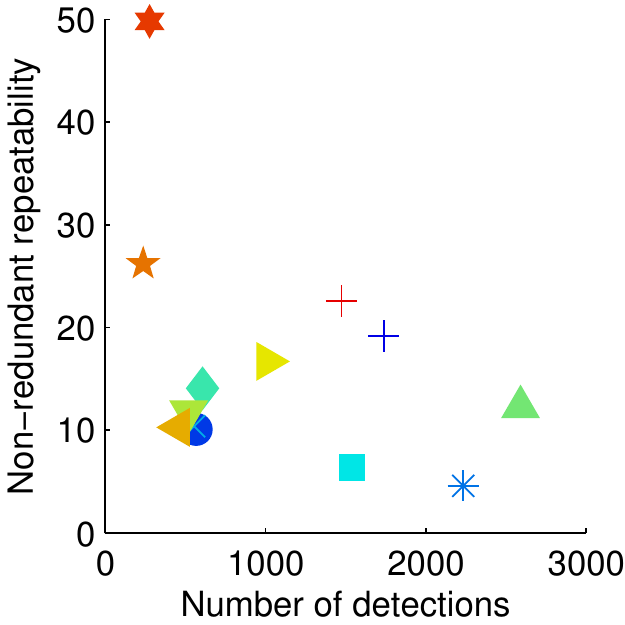}
        \end{minipage}
        \hspace{.6em}
        \begin{minipage}[c]{0.49\columnwidth}
            \centering
            \includegraphics[width=.8\textwidth]{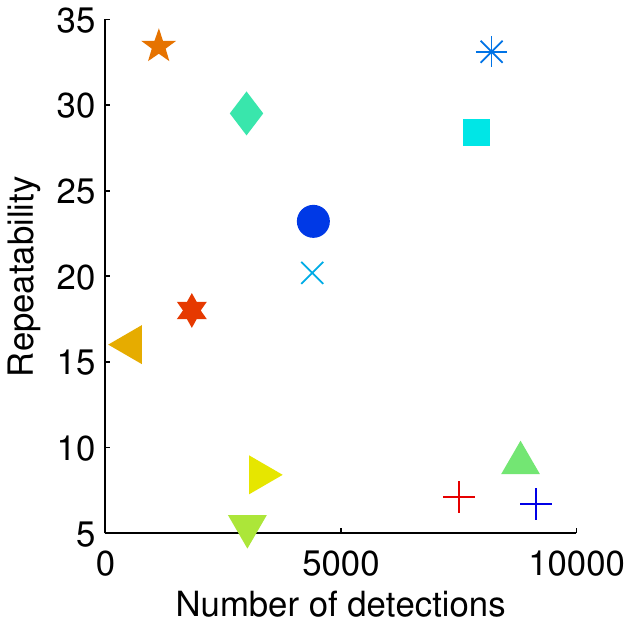}
        \end{minipage}
        \begin{minipage}[c]{0.49\columnwidth}
            \centering
            \includegraphics[width=.8\textwidth]{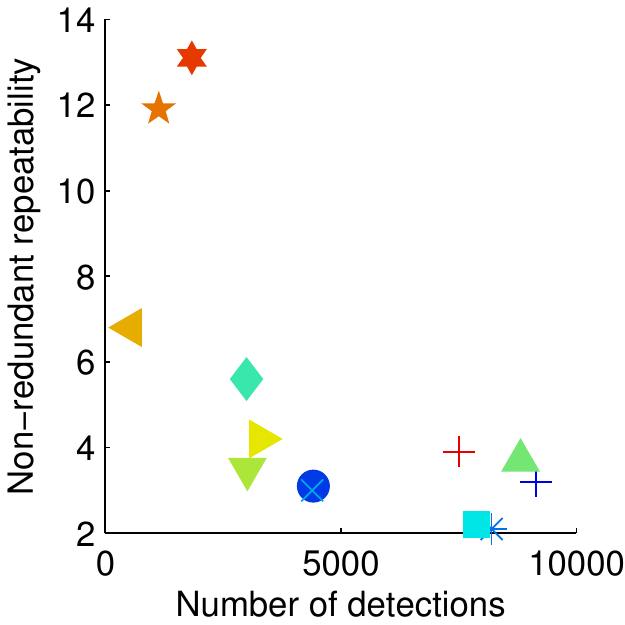}
        \end{minipage}

        \vspace{1.0em}
        \begin{minipage}[c]{0.98\columnwidth}
            \centering
            {\bf (e)} \texttt{leuven}  \emph{(illumination)}
        \end{minipage}
        \begin{minipage}[c]{0.98\columnwidth}
            \centering
            {\bf (f)} \texttt{trees}  \emph{(blur)}
        \end{minipage}
        \vspace{1.0em}

        \begin{minipage}[c]{0.49\columnwidth}
            \centering
            \includegraphics[width=.8\textwidth]{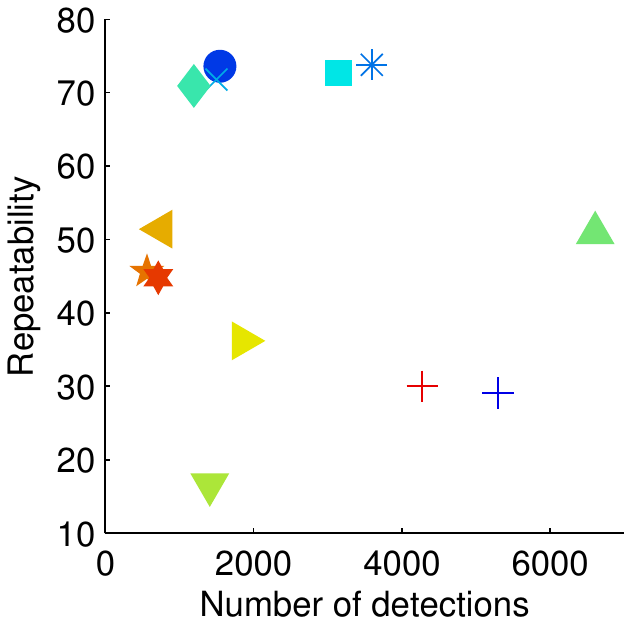}
        \end{minipage}
        \begin{minipage}[c]{0.49\columnwidth}
            \centering
            \includegraphics[width=.8\textwidth]{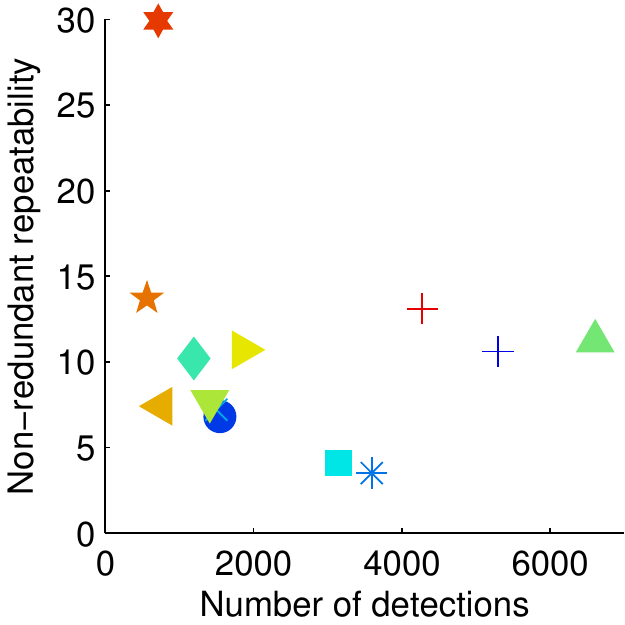}
        \end{minipage}
        \hspace{.6em}
        \begin{minipage}[c]{0.49\columnwidth}
            \centering
            \includegraphics[width=.8\textwidth]{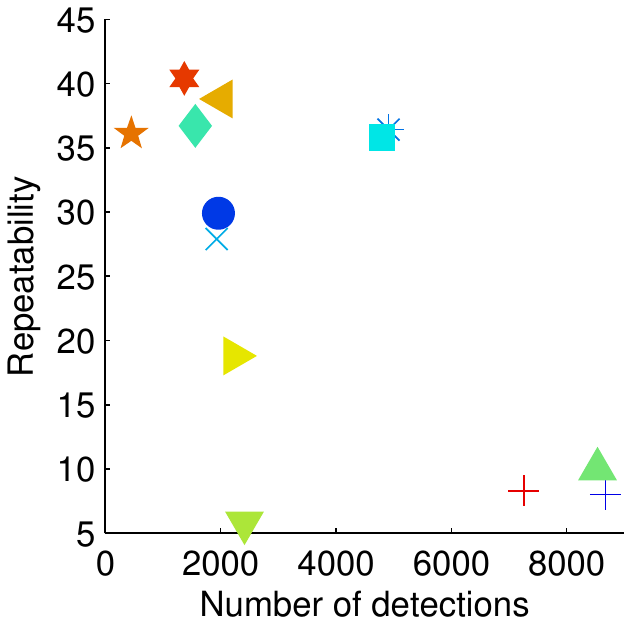}
        \end{minipage}
        \begin{minipage}[c]{0.49\columnwidth}
            \centering
            \includegraphics[width=.8\textwidth]{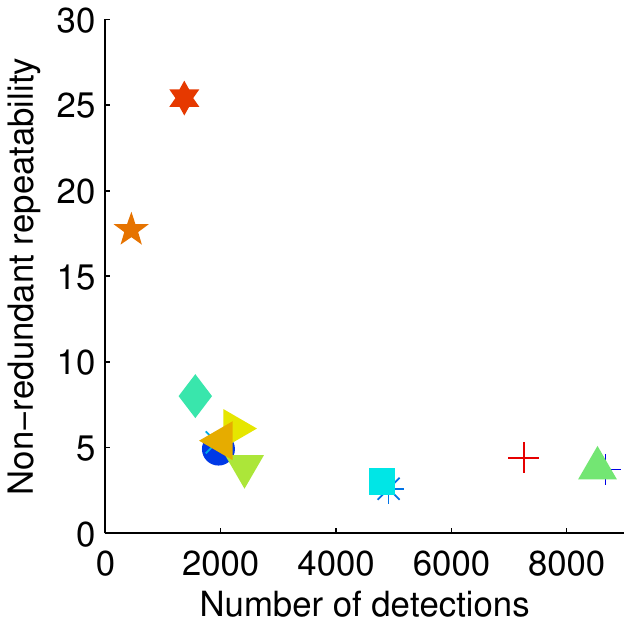}
        \end{minipage}

        \vspace{1.0em}
        \begin{minipage}[c]{0.98\columnwidth}
            \centering
            {\bf (g)} \texttt{ubc}  \emph{(jpeg)}
        \end{minipage}
        \begin{minipage}[c]{0.98\columnwidth}
            \centering
            {\bf (h)} \texttt{wall}  \emph{(viewpoint)}
        \end{minipage}

        \vspace{1.2em}
  \includegraphics[width=.5\textwidth]{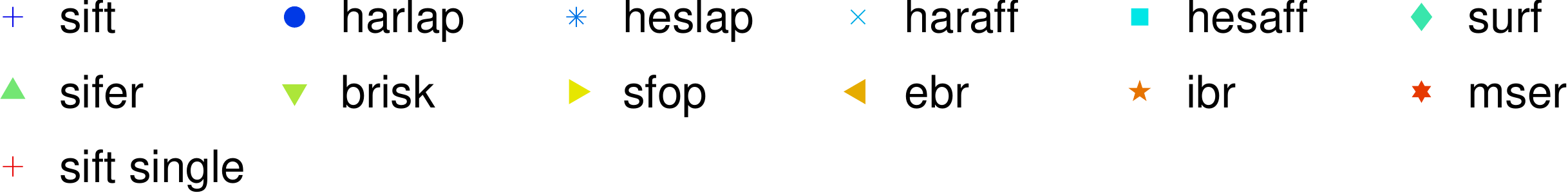}

    \end{center}
\caption{\small
    The average of the repeatability and non-redundant repeatability on each Oxford sequence
    is plotted as a function of the average number of keypoints detected.
    The performance evaluation of a detector is two-dimensional. On the one hand,
    a detector should detect as many keypoints as possible (abscissa).
    On the other, the detections should be as repeatable as possible (ordinate).
    Good detectors are on the top-right region of this plot.
   To compare a single detector performance the reader might follow the relative ordinate
    position of a particular detector in a particular scene in the traditional repeatability  (left)
    and the non-redundant repeatability plots (right).  For instance, MSER and SIFT algorithms always
    go up from the traditional to the non-redundant repeatability plots. This means that MSER and SIFT
    detections are less redundant than the average. On the other side, Hessian based methods and EBR/IBR
    always go down from the traditional to the non-redundant repeatability indicating redundant detections.
}
        \label{fig:mean:repeat}
\end{figure*}

The classic repeatability and the non-redundant repeatability rates
as well as the number of detections for the eight Oxford sequences
are provided in Table~\ref{tab:all}.
Also, in Figure~\ref{fig:mean:repeat} the average repeatability rates for the
13 compared detectors are plotted as a function of the number of detections.
Note that in general, the number of repeated points oscillates around $40\%$ of
the total number of detections. This is a much lower rate than usually achieved
with the more permissive definition of the repeatability criterion, see
Section~\ref{sec:repeat:bias}.

As previously said, the repeatability score must be compared alongside the number of
detections to have a complete performance evaluation of detectors.
The methods that provide in general the largest number of detections are SIFT,
SIFER and the Hessian based methods.
MSER, EBR and IBR produce significantly less detections.
The methods that are the most redundant happen to be also the
methods that perform well according to the classic repeatability criteria.
Indeed, the Hessian based methods are among the methods with  largest
repeatability while providing numerous detections.  Note that SFOP is
outperformed by the Harris based methods in all eight sequences, while providing
a similar number of detections.

These conclusions are drastically altered when the  redundancy of detections is taken into account.
According to the non-redundant repeatability shown in Table~\ref{tab:all}~{\bf (d)},
the hardly redundant SIFT method achieves one of the top three best scores while
providing in general one of the largest number of detections.
The Hessian based methods and SIFER, while achieving detection numbers
comparable to those of SIFT, perform poorly according to the non-redundant repeatability.
Despite having fewer detections, the non-redundant repeatability of SURF  is lower than the one of SIFT in  five sequences out of eight.
Unlike what was concluded with the classic criterion, SFOP outperforms the Harris based methods
in seven out of eight sequences. In fact, SFOP performs generally well. In all sequences, SFOP
is one of the three best algorithms according to the non-redundant repeatability while it performed poorly
for the traditional repeatability.
On average, MSER and IBR produce the best non-redundant repeatability scores.
Nevertheless, with up to ten times more detections, SIFT should be
preferred to MSER except for severe changes of viewpoint (see
Figure~\ref{fig:mean:repeat}).
In principle, MSER is not blur invariant. Yet, it performs
surprisingly well on the sequence \texttt{bikes}, containing well contrasted large geometric features.
MSER may benefit here from its low number of detections.

\begin{figure}[hptb]
    \begin{center}
       \small
        \begin{minipage}[c]{.45\columnwidth}
            \centering
            \includegraphics[width=\textwidth]{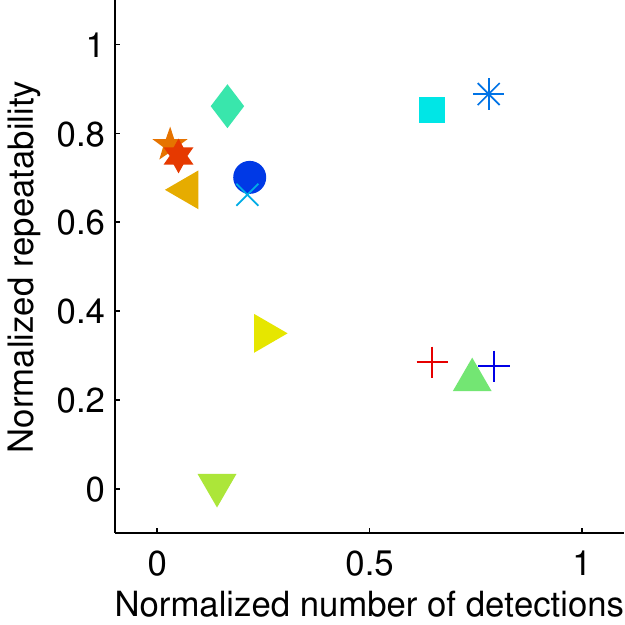}
        \end{minipage}
        \begin{minipage}[c]{.45\columnwidth}
            \centering
            \includegraphics[width=\textwidth]{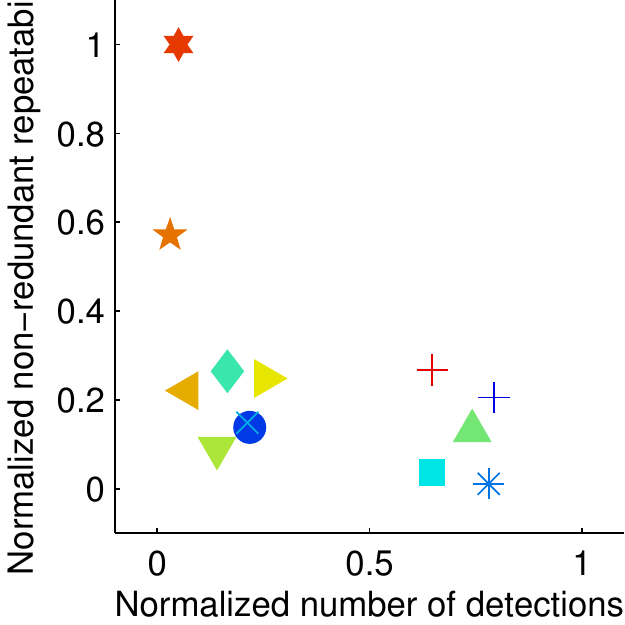}
        \end{minipage}

        \vspace{1em}
  \includegraphics[width=.97\columnwidth]{legend_mean4-crop.pdf}
    \end{center}
\caption{
    Qualitative visualization of the methods repeatability performance.
    For each sequence in the Oxford dataset, the number of detections,
    the repeatability and the non-redundant repeatability are scaled
    to the full range of $[0,1]$.
    Once normalized, the mean values of each method over the eight sequences are computed.
    On the left, the {\it rescaled} repeatability is plotted as
    a function of the {\it rescaled} number of detections. On the right, the
    {\it rescaled} non-redundant repeatability is plotted as a function of
    the {\it rescaled} number of detections.
    The same conclusions observed in each of the eight Oxford sequences apply
    is this qualitative contest.  
}
        \label{fig:rank:mean}
\end{figure}

To summarize the relative performance of each method on the entire Oxford data set we proceeded as follows.
First, the number of detections, the repeatability and non-redundant repeatability rates
on each sequence were rescaled to cover the interval $[0,1]$.
Then, we computed  the mean of the rescaled detectors performance over the eight sequences. Figure~\ref{fig:rank:mean} shows the
relative repeatability and non-redundant repeatability scores as a function of the  number of the normalized number detections.  
In this map a method performs optimally  if it is simultaneously extremal in ordinate and in abscissa, and performs well if it is extremal in at least one of the coordinates. 
Thus, the normalized benchmark reveals that the ranking of detectors is severely disrupted when considering the detectors redundancy.
While for example Harris and Hessian based methods, SURF and EBR
significantly reduce their performance (going down in the plot), MSER and BRISK improve their relative position to the others.
When the redundancy is not taken into account the method producing the most
detections and with the highest repeatability is Hessian Laplace, while when considering the non-redundant variant it is SIFT.

\setlength{\tabcolsep}{0.2em}
\begin{table*}[htpb]
\parbox{.5\linewidth}{
\scriptsize
\centering
 \begin{tabular}{lccccccccc}
    \toprule
        & \texttt{bark} & \texttt{bikes} & \texttt{boat} & \texttt{graf} & \texttt{leuven} & \texttt{trees} & \texttt{ubc} & \texttt{wall} & {\it mean} \\
        & {\it scale}   & {\it blur}     & {\it scale}   & {\it viewp}   & {\it illum}     & {\it blur}     & {\it jpeg}   & {\it viewp}   & \\
 SIFT   & \tbb{0.26}    & \tbb{0.27}     & \tbb{0.26}    & \tbb{0.29}    & \tbb{0.35}      & \tbb{0.20}     & \tbb{0.24}   & \tbb{0.23}    & {\it  \tbb{0.24}} \\
 SIFT-S & \tbb{0.31}    & \tbb{0.32}     & \tbb{0.31}    & \tbb{0.34}    & \tbb{0.41}      & \tbb{0.24}     & \tbb{0.29}   & \tbb{0.27}    & {\it  \tbb{0.29}} \\
 EBR    & \tb{0.23}     & \tb{0.18}      & \tb{0.10}     & \tb{0.11}     & \tb{0.15}       & \tbb{0.23}     & \tb{0.13}    & \tb{0.08}     & {\it  \tb{0.12}}  \\
 IBR    & \tb{0.20}     & \tb{0.24}      & \tb{0.21}     & \tb{0.21}     & \tbb{0.30}      & \tb{0.20}      & \tbb{0.21}   & \tbb{0.28}    & {\it  \tbb{0.22}} \\
 HARLAP & \tb{0.11}     & \tb{0.10}      & \tb{0.06}     & \tb{0.06}     & \tb{0.12}       & \tb{0.04}      & \tb{0.08}    & \tb{0.07}     & {\it  \tb{0.07}}  \\
 HESLAP & \tb{0.04}     & \tb{0.04}      & \tb{0.03}     & \tb{0.03}     & \tb{0.05}       & \tb{0.03}      & \tb{0.04}    & \tb{0.03}     & {\it  \tb{0.03}}  \\
 HARAFF & \tb{0.12}     & \tb{0.10}      & \tb{0.07}     & \tb{0.07}     & \tb{0.13}       & \tb{0.05}      & \tb{0.08}    & \tb{0.08}     & {\it  \tb{0.07}}  \\
 HESAFF & \tb{0.04}     & \tb{0.05}      & \tb{0.05}     & \tb{0.04}     & \tb{0.07}       & \tb{0.03}      & \tb{0.04}    & \tb{0.04}     & {\it  \tb{0.04}}  \\
 MSER   & \tba{0.61}    & \tba{0.55}     & \tba{0.51}    & \tba{0.55}    & \tba{0.58}      & \tba{0.48}     & \tba{0.55}   & \tba{0.48}    & {\it  \tba{0.51}} \\
 SURF   & \tb{0.16}     & \tb{0.16}      & \tb{0.11}     & \tb{0.11}     & \tb{0.16}       & \tb{0.10}      & \tb{0.13}    & \tb{0.13}     & {\it  \tb{0.12}}  \\
 SFOP   & \tb{0.17}     & \tb{0.24}      & \tbb{0.21}    & \tb{0.26}     & \tb{0.25}       & \tb{0.17}      & \tb{0.19}    & \tb{0.18}     & {\it  \tb{0.20}}  \\
 BRISK  & \tb{0.26}     & \tbb{0.28}     & \tb{0.14}     & \tbb{0.27}    & \tb{0.26}       & \tb{0.10}      & \tb{0.17}    & \tb{0.13}     & {\it  \tb{0.15}}  \\
 SIFER  & \tbb{0.31}    & \tb{0.23}      & \tb{0.18}     & \tb{0.21}     & \tb{0.22}       & \tb{0.16}      & \tb{0.18}    & \tb{0.19}     & {\it  \tb{0.19}}  \\
    \midrule
 \end{tabular}
 \vspace{0.5em}

\normalsize
{\bf (a)} Average {\it non-redundant ratio} $\text{nr} := K_\text{nr}/K$.
}
\parbox{.5\linewidth}{
\scriptsize
\centering
 \begin{tabular}{lccccccccc}
    \toprule
               & \texttt{bark} & \texttt{bikes} & \texttt{boat} & \texttt{graf} & \texttt{leuven} & \texttt{trees} & \texttt{ubc} & \texttt{wall} & {\it mean} \\
               & {\it scale}   & {\it blur}     & {\it scale}   & {\it viewp}   & {\it illum}     & {\it blur}     & {\it jpeg}   & {\it viewp}   & \\
        SIFT   & \tba{1021.6}  & \tbd{1034.8}   & \tbb{3802.8}  & \tbc{1906.6}  & \tbc{1736.6}    & \tba{9143.0}   & \tbb{5296.0} & \tba{8677.6}  & {\it \tbb{ 4077.4 }} \\
        SIFT-S & \tbb{848.0}   & \tb{871.0}     & \tbc{3225.6}  & \tbd{1641.8}  & \tb{1473.6}     & \tb{7506.8}    & \tbc{4272.6} & \tbc{7255.2}  & {\it \tbd{ 3386.8 }} \\
        EBR    & \tb{75.2}     & \tb{366.4}     & \tb{665.2}    & \tb{577.4}    & \tb{458.0}      & \tb{535.2}     & \tb{756.0}   & \tb{2012.4}   & {\it \tb{ 680.7 }} \\
        IBR    & \tb{131.8}    & \tb{573.2}     & \tb{280.6}    & \tb{293.6}    & \tb{238.2}      & \tb{1141.0}    & \tb{563.2}   & \tb{453.4}    & {\it \tb{ 459.4 }} \\
        HARLAP & \tb{118.0}    & \tb{541.0}     & \tb{1438.8}   & \tb{1120.8}   & \tb{568.4}      & \tb{4419.6}    & \tb{1549.0}  & \tb{1963.4}   & {\it \tb{ 1464.9 }} \\
        HESLAP & \tbc{814.6}   & \tba{2936.4}   & \tbd{2794.8}  & \tba{3164.8}  & \tbb{2233.2}    & \tbc{8201.6}   & \tbd{3594.0} & \tbd{4913.8}  & {\it \tbc{ 3581.7 }} \\
        HARAFF & \tb{120.2}    & \tb{533.2}     & \tb{1392.2}   & \tb{1103.0}   & \tb{555.8}      & \tb{4397.6}    & \tb{1501.0}  & \tb{1931.6}   & {\it \tb{ 1441.8 }} \\
        HESAFF & \tbd{807.2}   & \tbb{2470.0}   & \tb{2217.2}   & \tbb{2180.2}  & \tbd{1538.6}    & \tbd{7875.8}   & \tb{3146.0}  & \tb{4798.4}   & {\it \tb{ 3129.2 }} \\
        MSER   & \tb{85.4}     & \tb{195.2}     & \tb{592.4}    & \tb{280.4}    & \tb{276.4}      & \tb{1839.4}    & \tb{716.0}   & \tb{1372.8}   & {\it \tb{ 669.8 }} \\
        SURF   & \tb{183.0}    & \tb{546.6}     & \tb{948.2}    & \tb{913.4}    & \tb{607.8}      & \tb{3000.0}    & \tb{1194.0}  & \tb{1564.2}   & {\it \tb{ 1119.7 }} \\
        SFOP   & \tb{476.0}    & \tbd{1040.8}   & \tb{825.8}    & \tb{530.2}    & \tb{1014.0}     & \tb{3293.0}    & \tb{1859.4}  & \tb{2243.2}   & {\it \tb{ 1410.3 }} \\
        BRISK  & \tb{119.2}    & \tb{194.2}     & \tb{1149.6}   & \tb{374.0}    & \tb{521.4}      & \tb{3016.6}    & \tb{1408.8}  & \tb{2413.2}   & {\it \tb{ 1149.6 }} \\
        SIFER  & \tb{159.4}    & \tb{729.8}     & \tba{4321.4}  & \tb{1570.6}   & \tba{2591.4}    & \tbb{8818.2}   & \tba{6609.8} & \tbb{8535.2}  & {\it \tba{ 4167.0 }} \\

    \midrule
\end{tabular}
 \vspace{0.5em}

\normalsize
{\bf (b)}  Average number of detections in the common area.
}

\vspace{1.5em}

\parbox{.5\linewidth}{
\scriptsize
\centering
 \begin{tabular}{lccccccccc}
    \midrule
               & \texttt{bark} & \texttt{bikes} & \texttt{boat} & \texttt{graf} & \texttt{leuven} & \texttt{trees} & \texttt{ubc} & \texttt{wall} & {\it mean} \\
               & {\it scale}   & {\it blur}     & {\it scale}   & {\it viewp}   & {\it illum}     & {\it blur}     & {\it jpeg}   & {\it viewp}   & \\
        SIFT   & \tb{23.4}     & \tb{44.3}      & \tb{17.6}     & \tb{11.8}     & \tb{42.5}       & \tb{6.7}       & \tb{29.1}    & \tb{8.0}      & {\it \tb{ 22.9}} \\
        SIFT-S & \tb{23.3}     & \tb{44.6}      & \tb{18.1}     & \tb{11.9}     & \tb{43.5}       & \tb{7.1}       & \tb{30.0}    & \tb{8.3}      & {\it \tb{ 23.4}} \\
        EBR    & \tb{7.5}      & \tbd{66.6}     & \tba{53.5}    & \tbc{38.6}    & \tb{55.1}       & \tb{16.0}      & \tb{51.4}    & \tbb{38.8}    & {\it \tb{ 40.9}} \\
        IBR    & \tb{37.2}     & \tb{51.9}      & \tb{46.4}     & \tbb{50.6}    & \tb{58.1}       & \tba{33.4}     & \tb{45.6}    & \tb{36.1}     & {\it \tbd{ 44.9}} \\
        HARLAP & \tbd{52.5}    & \tb{52.4}      & \tb{40.2}     & \tb{21.3}     & \tb{50.2}       & \tbd{23.2}     & \tbb{73.6}   & \tb{29.9}     & {\it \tb{ 42.9}} \\
        HESLAP & \tbb{57.9}    & \tbb{69.5}     & \tbb{50.0}    & \tb{22.4}     & \tbb{70.1}      & \tbb{33.1}     & \tba{73.8}   & \tbd{36.4}    & {\it \tba{ 51.7}} \\
        HARAFF & \tb{48.6}     & \tb{50.0}      & \tb{36.7}     & \tb{26.9}     & \tb{47.5}       & \tb{20.2}      & \tbd{71.8}   & \tb{27.9}     & {\it \tb{ 41.2}} \\
        HESAFF & \tbc{54.7}    & \tbc{66.8}     & \tb{46.8}     & \tbd{30.7}    & \tbc{65.9}      & \tbc{28.4}     & \tbc{72.7}   & \tb{35.8}     & {\it \tbc{ 50.2}} \\
        MSER   & \tb{32.9}     & \tb{52.2}      & \tb{42.4}     & \tba{55.6}    & \tba{72.8}      & \tb{18.0}      & \tb{44.8}    & \tba{40.4}    & {\it \tb{ 44.9}} \\
        SURF   & \tba{63.6}    & \tba{72.6}     & \tbd{48.2}    & \tb{19.4}     & \tbd{64.6}      & \tb{29.5}      & \tb{70.9}    & \tbc{36.7}    & {\it \tbb{ 50.7}} \\
        SFOP   & \tb{29.7}     & \tb{31.8}      & \tb{25.9}     & \tb{13.7}     & \tb{42.6}       & \tb{8.4}       & \tb{36.2}    & \tb{18.8}     & {\it \tb{ 25.9}} \\
        BRISK  & \tb{2.4}      & \tb{9.9}       & \tb{4.0}      & \tb{4.3}      & \tb{18.2}       & \tb{5.4}       & \tb{16.6}    & \tb{5.8}      & {\it \tb{ 8.3}} \\
        SIFER  & \tb{1.4}      & \tb{49.9}      & \tb{7.4}      & \tb{1.5}      & \tb{37.5}       & \tb{9.1}       & \tb{50.9}    & \tb{10.0}     & {\it \tb{ 20.9}} \\

    \midrule
       \end{tabular}
       \vspace{0.5em}

    \normalsize
    {\bf (c)} Average repeatability.
}
\parbox{.5\linewidth}{
\scriptsize
\centering
 \begin{tabular}{lccccccccc}
    \toprule
               & \texttt{bark} & \texttt{bikes} & \texttt{boat} & \texttt{graf} & \texttt{leuven} & \texttt{trees} & \texttt{ubc} & \texttt{wall} & {\it mean} \\
               & {\it scale}   & {\it blur}     & {\it scale}   & {\it viewp}   & {\it illum}     & {\it blur}     & {\it jpeg}   & {\it viewp}   & \\
        SIFT   & \tb{7.3}      & \tbb{15.2}     & \tb{6.6}      & \tb{4.5}      & \tbb{19.2}      & \tb{3.2}       & \tb{10.6}    & \tb{3.7}      & {\it  \tb{8.8}}  \\
        SIFT-S & \tb{8.8}      & \tbb{18.1}     & \tbb{7.8}     & \tb{5.3}      & \tbb{22.6}      & \tb{3.9}       & \tbb{13.1}   & \tb{4.4}      & {\it  \tb{10.5}}  \\
        EBR    & \tb{5.3}      & \tbb{15.4}     & \tb{6.6}      & \tbb{9.2}     & \tb{10.3}       & \tbb{6.8}      & \tb{7.4}     & \tb{5.4}      & {\it  \tb{8.3}}  \\
        IBR    & \tbb{19.8}    & \tb{15.2}      & \tbb{15.4}    & \tbb{17.5}    & \tbb{26.2}      & \tbb{11.9}     & \tbb{13.7}   & \tbb{17.7}    & {\it  \tbb{17.2}} \\
        HARLAP & \tbb{11.1}    & \tb{9.1}       & \tb{4.1}      & \tb{2.6}      & \tb{10.1}       & \tb{3.1}       & \tb{6.8}     & \tb{4.9}      & {\it  \tb{6.5}}  \\
        HESLAP & \tb{3.8}      & \tb{3.7}       & \tb{2.4}      & \tb{1.2}      & \tb{4.6}        & \tb{2.1}       & \tb{3.5}     & \tb{2.6}      & {\it  \tb{3.0}}  \\
        HARAFF & \tb{11.1}     & \tb{9.4}       & \tb{4.0}      & \tb{4.2}      & \tb{10.4}       & \tb{3.0}       & \tb{7.2}     & \tb{5.3}      & {\it  \tb{6.8}}  \\
        HESAFF & \tb{4.1}      & \tb{4.6}       & \tb{2.8}      & \tb{2.8}      & \tb{6.4}        & \tb{2.2}       & \tb{4.1}     & \tb{3.0}      & {\it  \tb{3.7}}  \\
        MSER   & \tba{27.2}    & \tba{35.9}     & \tba{24.0}    & \tba{32.8}    & \tba{49.8}      & \tba{13.1}     & \tba{29.9}   & \tba{25.4}    & {\it  \tba{29.8}} \\
        SURF   & \tbb{14.7}    & \tb{13.3}      & \tb{7.1}      & \tb{3.7}      & \tb{14.1}       & \tbb{5.6}      & \tb{10.2}    & \tbb{8.0}     & {\it  \tbb{9.6}} \\
        SFOP   & \tb{10.0}     & \tb{11.5}      & \tbb{10.3}    & \tbb{6.2}     & \tb{16.7}       & \tb{4.2}       & \tb{10.7}    & \tbb{6.1}     & {\it  \tbb{9.5}} \\
        BRISK  & \tb{2.3}      & \tb{7.2}       & \tb{2.7}      & \tb{3.4}      & \tb{11.8}       & \tb{3.5}       & \tb{7.7}     & \tb{3.9}      & {\it  \tb{5.3}}  \\
        SIFER  & \tb{1.2}      & \tb{14.7}      & \tb{3.4}      & \tb{1.2}      & \tb{12.3}       & \tb{3.7}       & \tbb{11.2}   & \tb{3.8}      & {\it  \tb{6.4}}  \\

    \midrule
\end{tabular}
\vspace{0.5em}

\normalsize
{\bf (d)} Average non-redundant repeatability.}

\caption{} \vspace{-1em}

 \normalfont\sffamily\normalsize
Detectors comparison regarding repeatability and non-redundant
 repeatability rates on the eight sequences of the Oxford dataset. The
 algorithm with best number is colored in \tba{red} and the next three in \tbb{bordeaux}. Each table
 focuses on a single metric: the (non-redundant) repeatability or the number of detections.
 A fair comparison should consider both metrics simultaneously (see Figure~\ref{fig:mean:repeat}).

 \label{tab:all}
\end{table*}

\vspace{.5em}

\noindent {\bf Matching scenario.} We also explored the algorithms performance on a matching scenario.
For that purpose, we adopted the same protocol as in~\cite{Miko2005}.
Given a SIFT feature vector on one of the images (the reference image),
the distance to all the feature vectors of the other image is computed.
If the distance to the nearest neighbor is less than $60\%$ the distance to the second
nearest neighbor (i.e., a relative threshold on the distance), then we consider the pair of detections
as a matching (as proposed in~\cite{Lowe2004}).
Table~\ref{tab:match}~{\bf (b)} shows the average total number of
matches while Table~\ref{tab:match}~{\bf (c)} presents the number of correct
matches, namely those that are consistent with the ground truth.
Like in the repeatability criterion, one match is considered correct if the
overlap error between the two matched keypoints (elliptical regions) is inferior to $40\%$.
Table~\ref{tab:match}~{\bf (d)} gives the number of non-redundant correct matches while
the number of detections in the common area are given in Table~\ref{tab:match}~{\bf (a)}.
Due in part to their large number of detections, the Hessian based methods
achieve in general the largest number of correct matches.
In particular, in the \texttt{ubc} sequence, the Hessian-Laplace and
Hessian-Affine provide almost twice more correct matches
than SIFT on average.
However, this apparent advantage of the Hessian based methods fades
away once the detection redundancy is taken into account, as revealed by
the number of non-redundant correct matches.

SIFT and its single orientation variant achieve the largest number of non-redundant correct
matches in most sequences. Although SIFER produces on average the maximum number of non-redundant
correct matches on the whole data set, it performs poorly on two sequences (\texttt{graf} and \texttt{bark}).

\begin{figure}[hptb]
    \begin{center}
       \small
        \begin{minipage}[c]{.45\columnwidth}
            \centering
            \includegraphics[width=\textwidth]{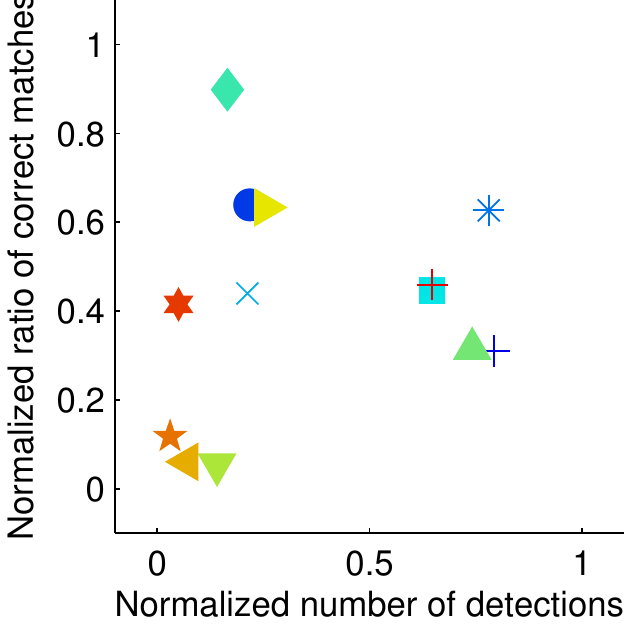}
        \end{minipage}
        \begin{minipage}[c]{.45\columnwidth}
            \centering
            \includegraphics[width=\textwidth]{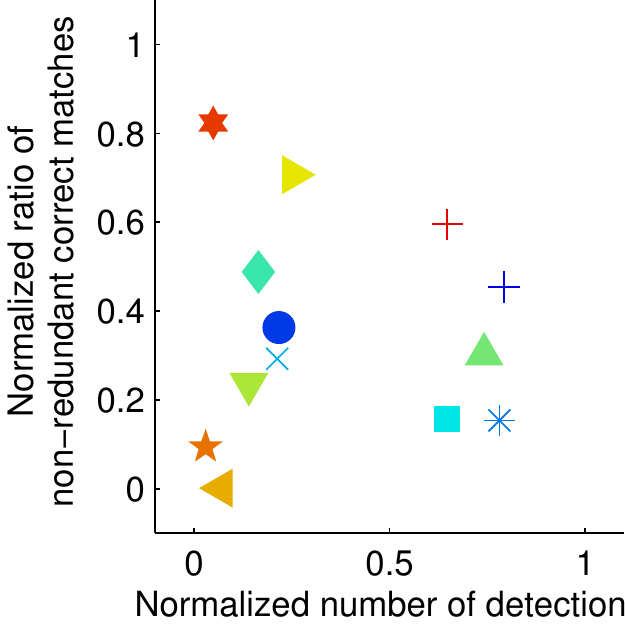}
        \end{minipage}

        \vspace{1em}
  \includegraphics[width=.97\columnwidth]{legend_mean4-crop.pdf}
    \end{center}
\caption{
    Qualitative visualization of the methods relative matching performances.
    For each sequence in the Oxford dataset, the number of detections,
    the ratio of correct matches and the ratio of non-redundant correct matches
    are rescaled in such a way as to range in $[0,1]$.
    In a matching scenario taking into account the redundancy of matches,
     SIFT  outperforms  Hessian based methods.
}
        \label{fig:rank:mean:matches}
\end{figure}

Figure~\ref{fig:rank:mean:matches} summarizes the methods matching performance
relatively to each other. For that purpose, the number of detections, the ratio of correct
matches and the ratio of non-redundant correct matches were
rescaled, and the mean values over the eight sequences of the rescaled ratios
are plotted as a function of the normalized number of detected keypoints.

Similarly to what we have observed on the repeatability ratio, the normalized matching benchmark reveals that the
ranking of detectors is significantly disrupted when considering the detectors redundancy.
Indeed, when the redundancy is not taken into account, the Hessian Laplace detector is the one producing more detections
and more number of correct matches per detection.  If instead we consider the redundancy, SIFT is the method
producing more detections and more non-redundant correct matches per detection.

Interestingly,  computing a single orientation for each
keypoint improves the performance of the SIFT method.  Indeed, this lowers the
computational cost of descriptor computations, increases the non-redundant
repeatability and maintains the number of non-redundant correct matches.

\setlength{\tabcolsep}{0.2em}
\begin{table*}[htpb]

\parbox{.5\linewidth}{
\scriptsize
\centering
 \begin{tabular}{lccccccccc}
    \toprule
               & \texttt{bark} & \texttt{bikes} & \texttt{boat} & \texttt{graf} & \texttt{leuven} & \texttt{trees} & \texttt{ubc} & \texttt{wall} & {\it mean} \\
               & {\it scale}   & {\it blur}     & {\it scale}   & {\it viewp}   & {\it illum}     & {\it blur}     & {\it jpeg}   & {\it viewp}   & \\
        SIFT   & \tba{1021.6}  & \tbd{1034.8}   & \tbb{3802.8}  & \tbc{1906.6}  & \tbc{1736.6}    & \tba{9143.0}   & \tbb{5296.0} & \tba{8677.6}  & {\it \tbb{ 4077.4 }} \\
        SIFT S & \tbb{848.0}   & \tb{871.0}     & \tbc{3225.6}  & \tbd{1641.8}  & \tb{1473.6}     & \tb{7506.8}    & \tbc{4272.6} & \tbc{7255.2}  & {\it \tbd{ 3386.8 }} \\
        EBR    & \tb{75.2}     & \tb{366.4}     & \tb{665.2}    & \tb{577.4}    & \tb{458.0}      & \tb{535.2}     & \tb{756.0}   & \tb{2012.4}   & {\it \tb{ 680.7 }} \\
        IBR    & \tb{131.8}    & \tb{573.2}     & \tb{280.6}    & \tb{293.6}    & \tb{238.2}      & \tb{1141.0}    & \tb{563.2}   & \tb{453.4}    & {\it \tb{ 459.4 }} \\
        HARLAP & \tb{118.0}    & \tb{541.0}     & \tb{1438.8}   & \tb{1120.8}   & \tb{568.4}      & \tb{4419.6}    & \tb{1549.0}  & \tb{1963.4}   & {\it \tb{ 1464.9 }} \\
        HESLAP & \tbc{814.6}   & \tba{2936.4}   & \tbd{2794.8}  & \tba{3164.8}  & \tbb{2233.2}    & \tbc{8201.6}   & \tbd{3594.0} & \tbd{4913.8}  & {\it \tbc{ 3581.7 }} \\
        HARAFF & \tb{120.2}    & \tb{533.2}     & \tb{1392.2}   & \tb{1103.0}   & \tb{555.8}      & \tb{4397.6}    & \tb{1501.0}  & \tb{1931.6}   & {\it \tb{ 1441.8 }} \\
        HESAFF & \tbd{807.2}   & \tbb{2470.0}   & \tb{2217.2}   & \tbb{2180.2}  & \tbd{1538.6}    & \tbd{7875.8}   & \tb{3146.0}  & \tb{4798.4}   & {\it \tb{ 3129.2 }} \\
        MSER   & \tb{85.4}     & \tb{195.2}     & \tb{592.4}    & \tb{280.4}    & \tb{276.4}      & \tb{1839.4}    & \tb{716.0}   & \tb{1372.8}   & {\it \tb{ 669.8 }} \\
        SURF   & \tb{183.0}    & \tb{546.6}     & \tb{948.2}    & \tb{913.4}    & \tb{607.8}      & \tb{3000.0}    & \tb{1194.0}  & \tb{1564.2}   & {\it \tb{ 1119.7 }} \\
        SFOP   & \tb{476.0}    & \tbd{1040.8}   & \tb{825.8}    & \tb{530.2}    & \tb{1014.0}     & \tb{3293.0}    & \tb{1859.4}  & \tb{2243.2}   & {\it \tb{ 1410.3 }} \\
        BRISK  & \tb{119.2}    & \tb{194.2}     & \tb{1149.6}   & \tb{374.0}    & \tb{521.4}      & \tb{3016.6}    & \tb{1408.8}  & \tb{2413.2}   & {\it \tb{ 1149.6 }} \\
        SIFER  & \tb{159.4}    & \tb{729.8}     & \tba{4321.4}  & \tb{1570.6}   & \tba{2591.4}    & \tbb{8818.2}   & \tba{6609.8} & \tbb{8535.2}  & {\it \tba{ 4167.0 }} \\
    \midrule
\end{tabular}
 \vspace{0.5em}

\normalsize
{\bf (a)}  Average number of detections in the common area.
}
\parbox{.5\linewidth}{
\scriptsize
\centering
 \begin{tabular}{lccccccccc}
\toprule
        & \texttt{bark} & \texttt{bikes} & \texttt{boat} & \texttt{graf} & \texttt{leuven} & \texttt{trees} & \texttt{ubc} & \texttt{wall} & {\it mean} \\
        & {\it scale}   & {\it blur}     & {\it scale}   & {\it viewp}   & {\it illum}     & {\it blur}     & {\it jpeg}   & {\it viewp}   & \\
 SIFT   & \tbb{328.0}   & \tb{271.6}     & \tbb{567.0}   & \tbb{170.6}   & \tbb{518.2}     & \tb{335.0}     & \tb{889.6}   & \tbb{1189.6}  & {\it  \tbb{533.7}} \\
 SIFT-S & \tba{388.4}   & \tbb{322.4}    & \tba{637.8}   & \tba{198.6}   & \tbb{585.4}     & \tb{370.4}     & \tbb{1040.0} & \tbb{732.4}   & {\it  \tbb{534.4}} \\
 EBR    & \tb{5.0}      & \tb{60.0}      & \tb{19.8}     & \tb{14.2}     & \tb{51.8}       & \tb{15.0}      & \tb{186.0}   & \tb{0.0}      & {\it  \tb{44.0}}  \\
 IBR    & \tb{9.2}      & \tb{69.8}      & \tb{13.2}     & \tb{13.4}     & \tb{30.0}       & \tb{48.8}      & \tb{112.2}   & \tb{19.8}     & {\it  \tb{39.6}}  \\
 HARLAP & \tb{21.4}     & \tb{203.0}     & \tb{244.4}    & \tb{53.2}     & \tb{154.0}      & \tb{338.6}     & \tb{943.2}   & \tb{210.4}    & {\it  \tb{271.0}}  \\
 HESLAP & \tbb{168.2}   & \tba{1125.2}   & \tbb{378.4}   & \tbb{124.8}   & \tbb{653.6}     & \tba{705.0}    & \tbb{2022.8} & \tbb{572.6}   & {\it  \tbb{718.8}} \\
 HARAFF & \tb{9.4}      & \tb{155.8}     & \tb{125.0}    & \tb{48.8}     & \tb{122.8}      & \tb{225.6}     & \tb{840.2}   & \tb{201.6}    & {\it  \tb{216.2}}  \\
 HESAFF & \tb{50.2}     & \tbb{857.4}    & \tb{148.4}    & \tbb{67.6}    & \tb{400.2}      & \tbb{507.6}    & \tbb{1636.0} & \tb{567.2}    & {\it  \tb{529.3}}  \\
 MSER   & \tb{7.2}      & \tb{67.0}      & \tb{37.8}     & \tb{12.0}     & \tb{108.6}      & \tb{61.2}      & \tb{194.6}   & \tb{154.6}    & {\it  \tb{80.4}}  \\
 SURF   & \tb{47.2}     & \tb{311.0}     & \tb{177.8}    & \tb{54.4}     & \tb{233.4}      & \tbb{410.2}    & \tb{741.2}   & \tb{261.4}    & {\it  \tb{279.6}}  \\
 SFOP   & \tbb{132.8}   & \tb{310.2}     & \tb{217.8}    & \tb{64.0}     & \tb{357.0}      & \tb{186.2}     & \tb{587.8}   & \tb{384.2}    & {\it  \tb{280.0}}  \\
 BRISK  & \tb{5.2}      & \tb{29.0}      & \tb{67.6}     & \tb{20.2}     & \tb{114.8}      & \tb{125.8}     & \tb{344.8}   & \tb{160.2}    & {\it  \tb{108.5}}  \\
 SIFER  & \tb{8.6}      & \tbb{313.0}    & \tbb{384.0}   & \tb{55.4}     & \tba{872.6}     & \tbb{553.0}    & \tba{2329.6} & \tba{1694.4}  & {\it  \tba{776.3}} \\
\midrule
\end{tabular}
 \vspace{0.5em}

\normalsize
{\bf (b)}  Total number of matches.
}

\vspace{1.5em}
\parbox{.5\linewidth}{
\scriptsize
\centering
 \begin{tabular}{lccccccccc}
   \toprule
        & \texttt{bark} & \texttt{bikes} & \texttt{boat} & \texttt{graf} & \texttt{leuven} & \texttt{trees} & \texttt{ubc} & \texttt{wall} & {\it mean} \\
        & {\it scale}   & {\it blur}     & {\it scale}   & {\it viewp}   & {\it illum}     & {\it blur}     & {\it jpeg}   & {\it viewp}   & \\
 SIFT   & \tbb{106.8}   & \tb{240.8}     & \tbb{365.2}   & \tbb{104.8}   & \tbb{420.2}     & \tb{161.8}     & \tb{758.6}   & \tbb{303.0}   & {\it  \tbb{307.7}} \\
 SIFT-S & \tbb{133.0}   & \tb{286.0}     & \tba{413.0}   & \tba{124.2}   & \tbb{474.6}     & \tb{180.6}     & \tb{894.4}   & \tb{128.4}    & {\it  \tb{329.3}}  \\
 EBR    & \tb{0.4}      & \tb{56.6}      & \tb{13.0}     & \tb{6.8}      & \tb{43.8}       & \tb{7.0}       & \tb{174.4}   & \tb{0.0}      & {\it  \tb{37.8}}  \\
 IBR    & \tb{2.2}      & \tb{64.6}      & \tb{8.4}      & \tb{5.8}      & \tb{27.2}       & \tb{30.2}      & \tb{102.4}   & \tb{14.2}     & {\it  \tb{31.9}}  \\
 HARLAP & \tb{18.8}     & \tb{189.2}     & \tbb{219.6}   & \tb{48.0}     & \tb{141.2}      & \tb{258.0}     & \tbb{927.6}  & \tb{173.6}    & {\it  \tb{247.0}}  \\
 HESLAP & \tba{138.2}   & \tba{1047.8}   & \tbb{326.8}   & \tbb{101.8}   & \tbb{609.8}     & \tba{536.8}    & \tbb{1942.0} & \tbb{456.4}   & {\it  \tbb{645.0}} \\
 HARAFF & \tb{7.8}      & \tb{142.6}     & \tb{102.8}    & \tb{42.2}     & \tb{109.2}      & \tb{172.2}     & \tb{819.6}   & \tb{158.6}    & {\it  \tb{194.4}}  \\
 HESAFF & \tb{41.2}     & \tbb{782.2}    & \tb{123.0}    & \tbb{49.6}    & \tb{366.4}      & \tbb{372.2}    & \tbb{1585.4} & \tbb{443.4}   & {\it  \tbb{470.4}} \\
 MSER   & \tb{4.6}      & \tb{65.6}      & \tb{32.0}     & \tb{8.0}      & \tb{106.2}      & \tb{50.6}      & \tb{189.8}   & \tb{133.8}    & {\it  \tb{73.8}}  \\
 SURF   & \tb{41.0}     & \tbb{294.2}    & \tb{157.2}    & \tb{45.0}     & \tb{211.2}      & \tbb{312.4}    & \tb{694.6}   & \tb{227.6}    & {\it  \tb{247.9}}  \\
 SFOP   & \tbb{76.4}    & \tb{248.8}     & \tb{179.8}    & \tb{47.8}     & \tb{295.6}      & \tb{116.0}     & \tb{532.2}   & \tb{241.8}    & {\it  \tb{217.3}}  \\
 BRISK  & \tb{2.0}      & \tb{13.6}      & \tb{28.2}     & \tb{8.2}      & \tb{57.0}       & \tb{47.8}      & \tb{176.8}   & \tb{51.2}     & {\it  \tb{48.1}}  \\
 SIFER  & \tb{0.8}      & \tbb{286.2}    & \tb{136.4}    & \tb{9.8}      & \tba{703.6}     & \tbb{263.0}    & \tba{2195.8} & \tba{504.2}   & {\it  \tba{512.5}} \\
    \midrule	
\end{tabular}
\vspace{0.5em}

\normalsize
{\bf (c)} Number of correct matches.

}
\parbox{.5\linewidth}{
\scriptsize
\centering
\begin{tabular}{lccccccccc}
    \toprule
         & \texttt{bark} & \texttt{bikes} & \texttt{boat} & \texttt{graf} & \texttt{leuven} & \texttt{trees} & \texttt{ubc} & \texttt{wall} & {\it mean} \\
         & {\it scale}   & {\it blur}     & {\it scale}   & {\it viewp}   & {\it illum}     & {\it blur}     & {\it jpeg}   & {\it viewp}   & \\
  SIFT   & \tbb{47.1}    & \tbb{106.7}    & \tbb{173.3}   & \tbb{52.8}    & \tbb{234.3}     & \tbb{91.0}     & \tbb{344.1}  & \tbb{181.3}   & {\it  \tbb{153.8}} \\
  SIFT-S & \tba{52.5}    & \tba{118.8}    & \tba{190.4}   & \tba{59.5}    & \tba{264.6}     & \tbb{101.2}    & \tbb{387.4}  & \tb{70.4}     & {\it  \tb{155.6}}  \\
  EBR    & \tb{0.0}      & \tb{6.9}       & \tb{2.3}      & \tb{1.5}      & \tb{5.4}        & \tb{2.3}       & \tb{7.9}     & \tb{0.0}      & {\it  \tb{3.3}}  \\
  IBR    & \tb{0.5}      & \tb{9.6}       & \tb{2.3}      & \tb{1.1}      & \tb{6.8}        & \tb{7.5}       & \tb{9.3}     & \tb{12.5}     & {\it  \tb{6.2}}  \\
  HARLAP & \tb{7.1}      & \tb{38.0}      & \tb{36.7}     & \tb{11.1}     & \tb{39.1}       & \tb{59.2}      & \tb{89.0}    & \tb{47.8}     & {\it  \tb{41.0}}  \\
  HESLAP & \tbb{19.2}    & \tb{80.5}      & \tb{39.5}     & \tb{13.9}     & \tb{74.9}       & \tbb{78.3}     & \tb{93.0}    & \tb{64.4}     & {\it  \tb{58.0}}  \\
  HARAFF & \tb{3.1}      & \tb{36.4}      & \tb{29.4}     & \tb{12.5}     & \tb{36.8}       & \tb{51.2}      & \tb{89.6}    & \tb{51.1}     & {\it  \tb{38.8}}  \\
  HESAFF & \tb{12.3}     & \tb{83.7}      & \tb{29.5}     & \tbb{14.2}    & \tb{70.3}       & \tb{69.7}      & \tb{98.8}    & \tb{72.2}     & {\it  \tb{56.4}}  \\
  MSER   & \tb{2.2}      & \tb{43.2}      & \tb{26.1}     & \tb{6.3}      & \tb{81.3}       & \tb{39.5}      & \tb{129.5}   & \tbb{106.2}   & {\it  \tbb{54.3}} \\
  SURF   & \tb{11.0}     & \tb{47.9}      & \tb{28.8}     & \tb{11.2}     & \tb{49.0}       & \tb{63.9}      & \tb{71.8}    & \tb{68.9}     & {\it  \tb{44.0}}  \\
  SFOP   & \tbb{31.2}    & \tbb{96.9}     & \tbb{70.2}    & \tbb{21.6}    & \tbb{130.4}     & \tb{62.6}      & \tbb{160.8}  & \tbb{96.3}    & {\it  \tbb{83.8}} \\
  BRISK  & \tb{0.8}      & \tb{8.5}       & \tb{19.8}     & \tb{6.2}      & \tb{39.1}       & \tb{32.5}      & \tb{81.2}    & \tb{39.2}     & {\it  \tb{28.4}}  \\
  SIFER  & \tb{0.2}      & \tbb{91.0}     & \tbb{73.7}    & \tb{7.5}      & \tbb{253.1}     & \tba{109.7}    & \tba{536.8}  & \tba{210.8}   & {\it  \tba{160.4}} \\
    \midrule
        \end{tabular}
        \vspace{0.5em}

\normalsize
{\bf (d)}  Number of non-redundant correct matches.
}

\caption{}
\label{tab:match}
\vspace{-1em}

 \normalfont\sffamily\normalsize
   The matching performance of the compared detectors on the eight sequences of the Oxford dataset.
    In \tba{red} the algorithm with the largest number in the column. The other top three are in \tbb{bordeaux}.
    The best algorithms is the one that produces the largest number of correct (non redundant) matches, provided it does not make too many detections.
    This is a bi-dimensional criterion that is not fully represented in a single table.  Another comparison will consider both components simultaneously (Figure~\ref{fig:matching:detections}).

\end{table*}

\begin{figure*}[hptb]
    \begin{center}
       \small
        \begin{minipage}[c]{0.49\columnwidth}
            \centering
            \includegraphics[width=.8\textwidth]{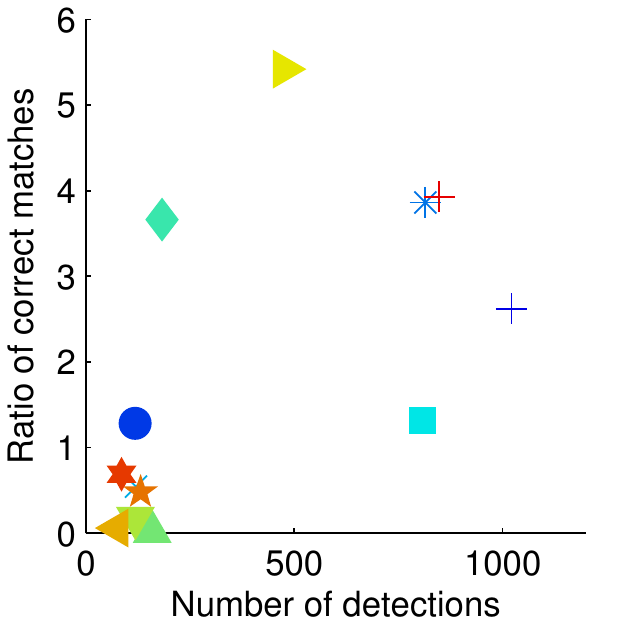}
        \end{minipage}
        \begin{minipage}[c]{0.49\columnwidth}
            \centering
            \includegraphics[width=.8\textwidth]{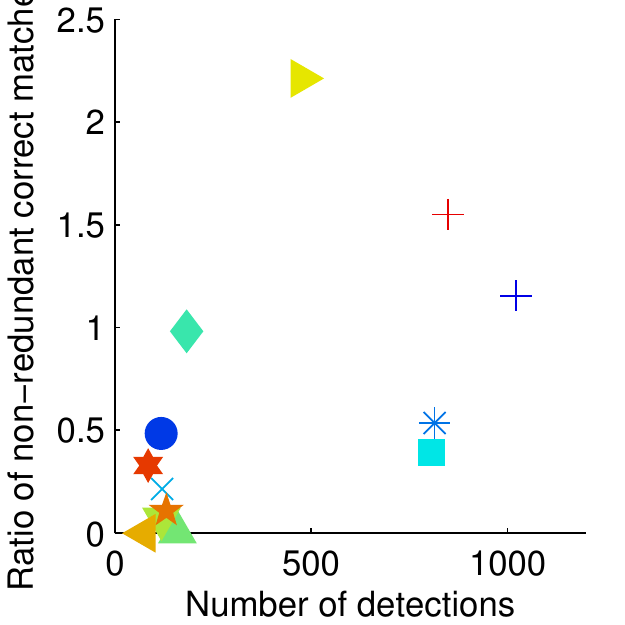}
        \end{minipage}
        \hspace{.6em}
        \begin{minipage}[c]{0.49\columnwidth}
            \centering
            \includegraphics[width=.8\textwidth]{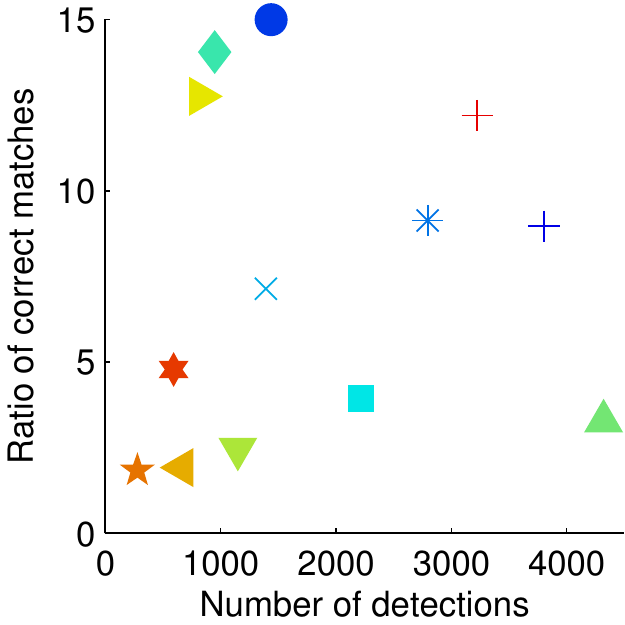}
        \end{minipage}
        \begin{minipage}[c]{0.49\columnwidth}
            \centering
            \includegraphics[width=.8\textwidth]{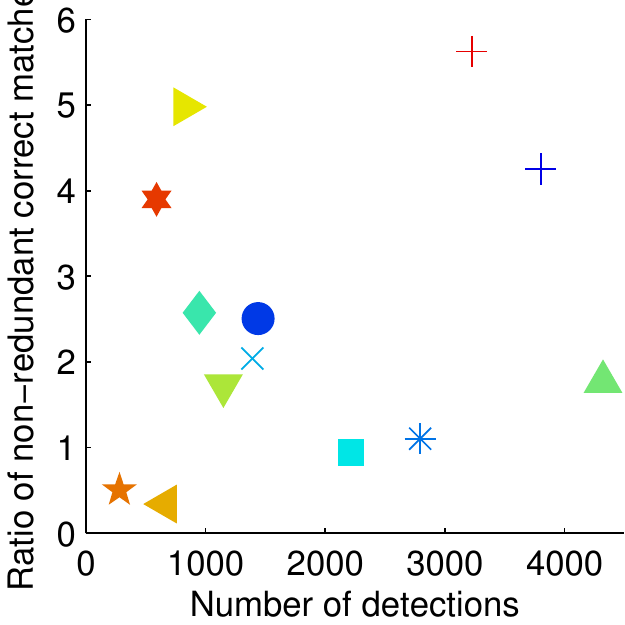}
        \end{minipage}

        \vspace{1.0em}
        \begin{minipage}[c]{0.98\columnwidth}
            \centering
            {\bf (a)} \texttt{bark}  \emph{(scale)   }
        \end{minipage}
        \begin{minipage}[c]{0.98\columnwidth}
            \centering
            {\bf (b)}  \texttt{boat} \emph{(scale) }
        \end{minipage}
        \vspace{1.0em}

        \begin{minipage}[c]{0.49\columnwidth}
            \centering
            \includegraphics[width=.8\textwidth]{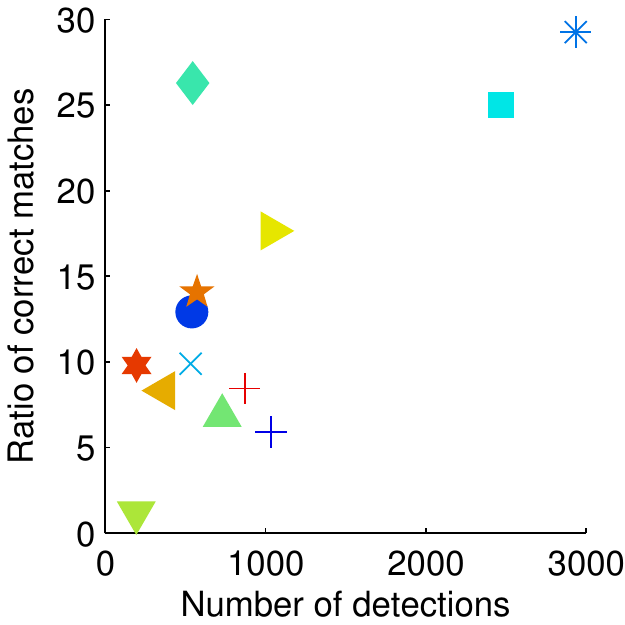}
        \end{minipage}
        \begin{minipage}[c]{0.49\columnwidth}
            \centering
            \includegraphics[width=.8\textwidth]{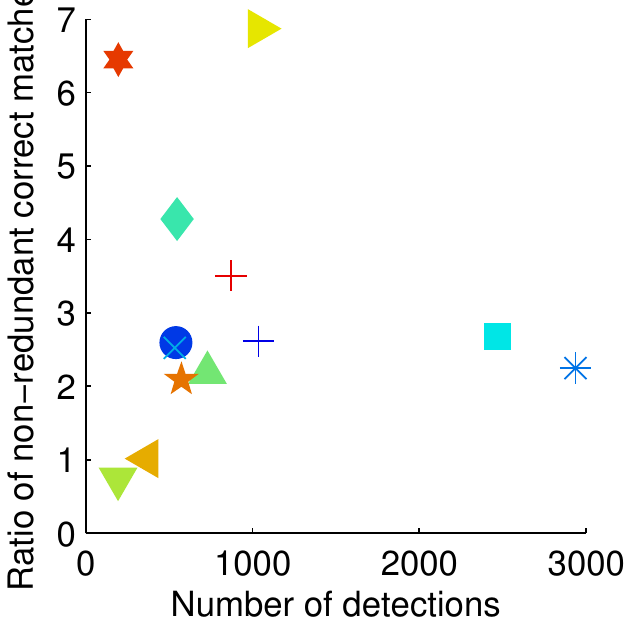}
        \end{minipage}
        \hspace{.6em}
        \begin{minipage}[c]{0.49\columnwidth}
            \centering
            \includegraphics[width=.85\textwidth]{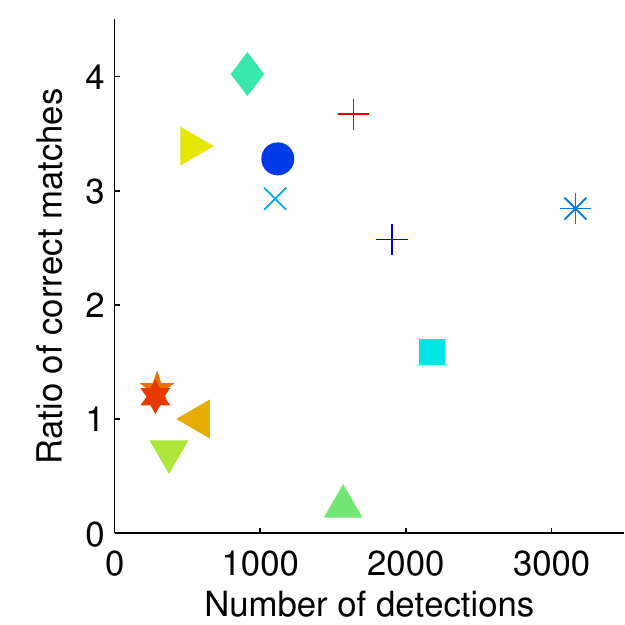}
        \end{minipage}
        \begin{minipage}[c]{0.49\columnwidth}
            \centering
            \includegraphics[width=.8\textwidth]{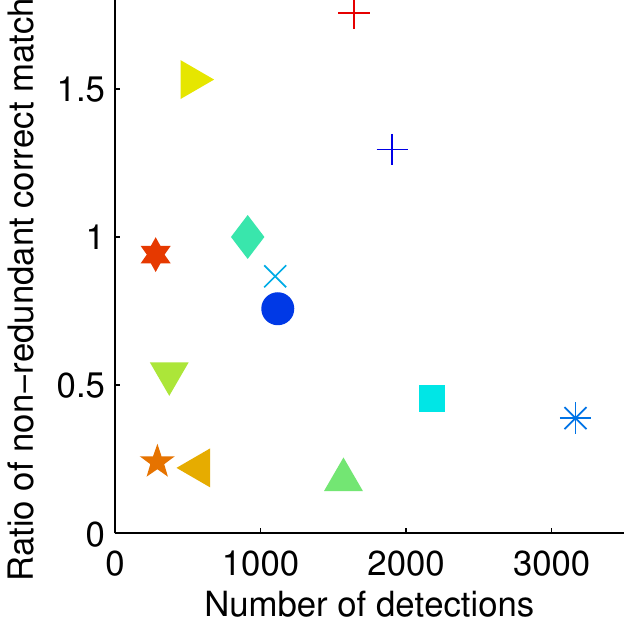}
        \end{minipage}

        \vspace{1.0em}
        \begin{minipage}[c]{0.98\columnwidth}
            \centering
            {\bf (c)}  \texttt{bikes}  \emph{(blur) }
        \end{minipage}
        \begin{minipage}[c]{0.98\columnwidth}
            \centering
            {\bf (d)}   \texttt{graf}  \emph{(viewpoint) }
        \end{minipage}
        \vspace{1.0em}

        \begin{minipage}[c]{0.49\columnwidth}
            \centering
            \includegraphics[width=.8\textwidth]{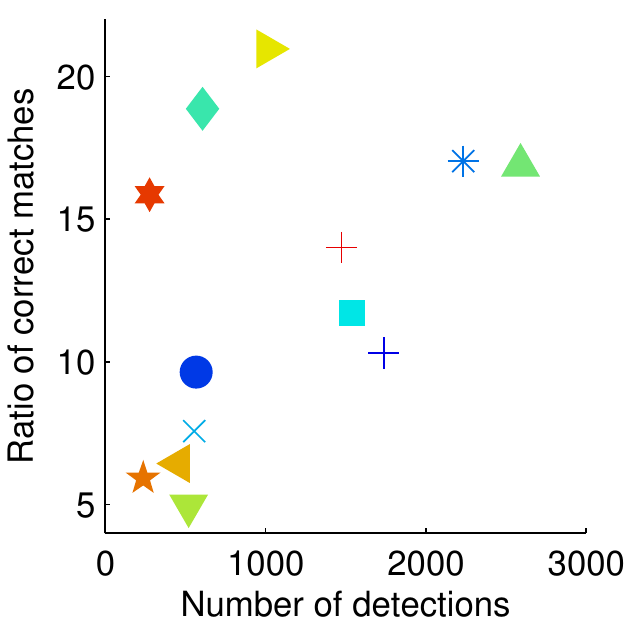}
        \end{minipage}
        \begin{minipage}[c]{0.49\columnwidth}
            \centering
            \includegraphics[width=.8\textwidth]{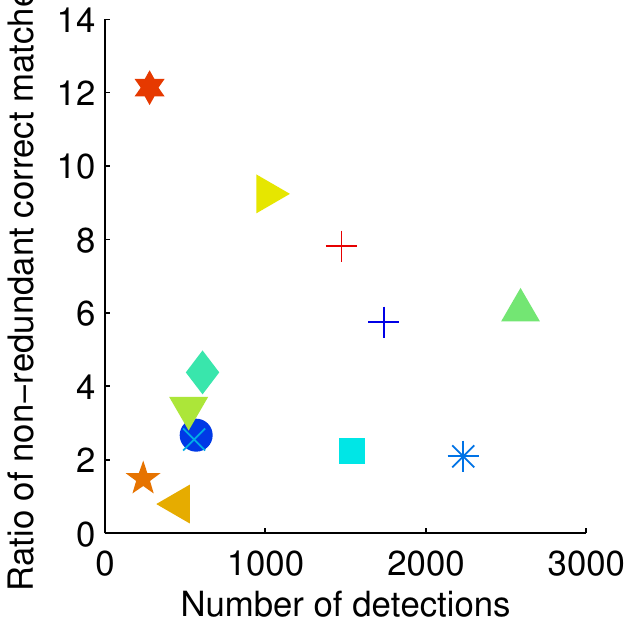}
        \end{minipage}
        \hspace{.6em}
        \begin{minipage}[c]{0.49\columnwidth}
            \centering
            \includegraphics[width=.8\textwidth]{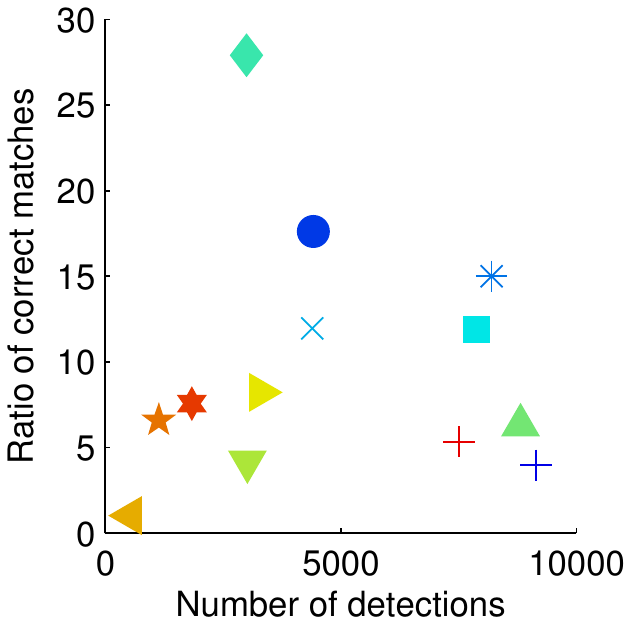}
        \end{minipage}
        \begin{minipage}[c]{0.49\columnwidth}
            \centering
            \includegraphics[width=.8\textwidth]{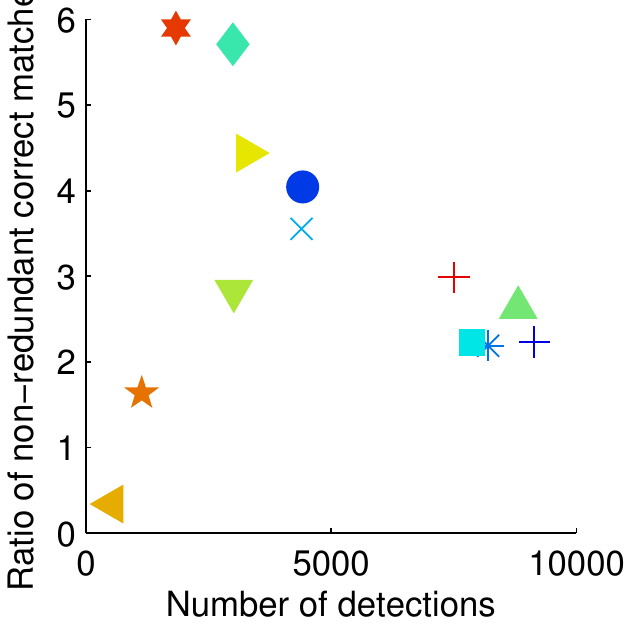}
        \end{minipage}

        \vspace{1.0em}
        \begin{minipage}[c]{0.98\columnwidth}
            \centering
            {\bf (e)}  \texttt{leuven}  \emph{(illumination) }
        \end{minipage}
        \begin{minipage}[c]{0.98\columnwidth}
            \centering
            {\bf (f)}   \texttt{trees}  \emph{(blur)}
        \end{minipage}
        \vspace{1.0em}

        \begin{minipage}[c]{0.49\columnwidth}
            \centering
            \includegraphics[width=.8\textwidth]{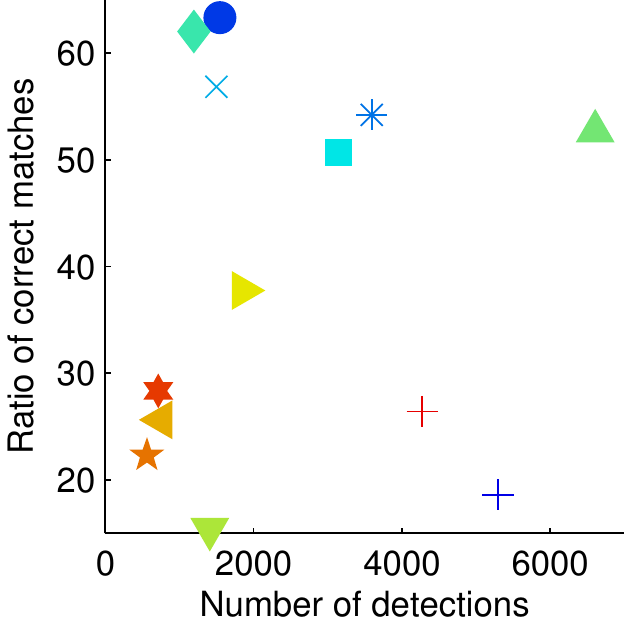}
        \end{minipage}
        \begin{minipage}[c]{0.49\columnwidth}
            \centering
            \includegraphics[width=.8\textwidth]{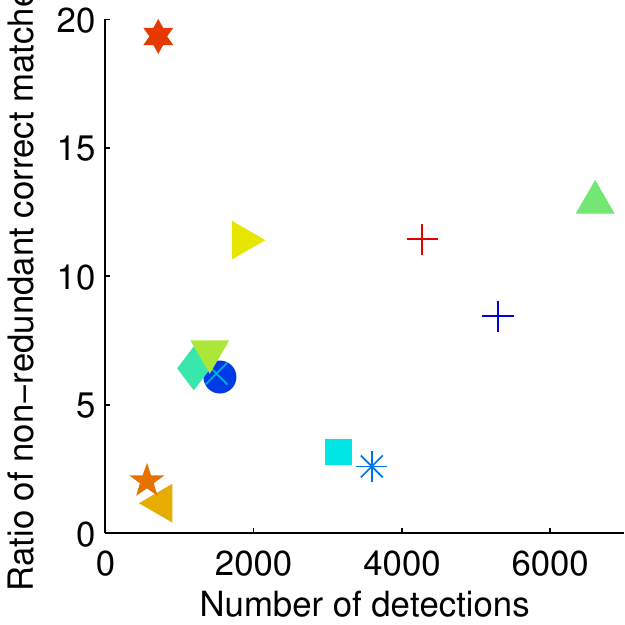}
        \end{minipage}
        \hspace{.6em}
        \begin{minipage}[c]{0.49\columnwidth}
            \centering
            \includegraphics[width=.8\textwidth]{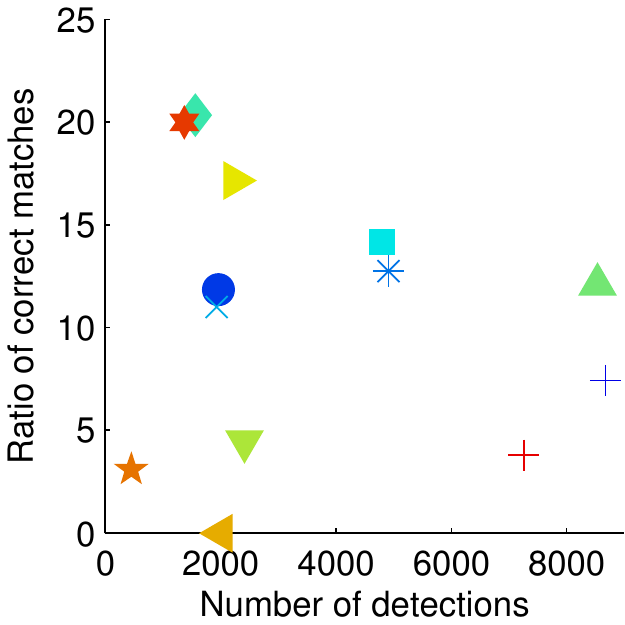}
        \end{minipage}
        \begin{minipage}[c]{0.49\columnwidth}
            \centering
            \includegraphics[width=.8\textwidth]{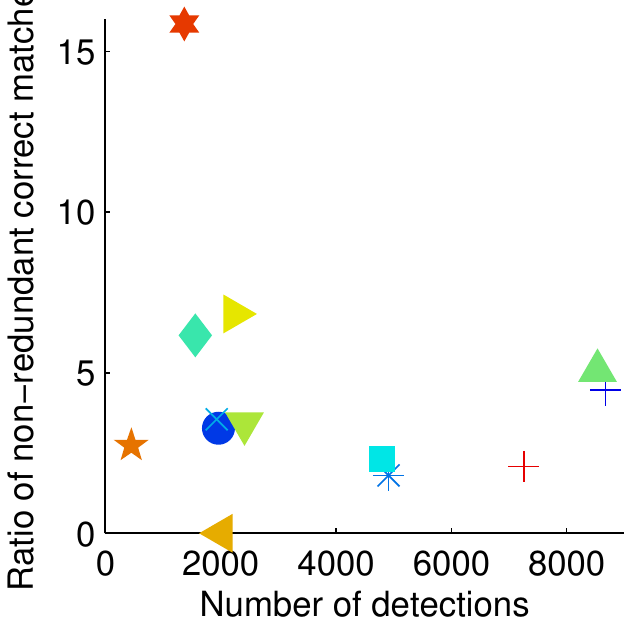}
        \end{minipage}

        \vspace{1.0em}
        \begin{minipage}[c]{0.98\columnwidth}
            \centering
            {\bf (g)} \texttt{ubc}  \emph{(jpeg)}
        \end{minipage}
        \begin{minipage}[c]{0.98\columnwidth}
            \centering
            {\bf (h)} \texttt{wall}  \emph{(viewpoint)}
        \end{minipage}

        \vspace{1.2em}

  \includegraphics[width=.5\textwidth]{legend_mean4-crop.pdf}
    \end{center}
\caption{
    Ratio of correct matches (left) and non-redundant correct matches (right) i.e., the
number of matches over number of detections in the area covered by both
images.
    Again, to compare a single detector matching performance the reader might
    follow the relative ordinate position of a particular detector in a
    particular scene.
    Generally, MSER, SIFT and SFOP algorithms go up once the redundancy of
    matches is taken into account.  On the other side, Hessian based methods
    and EBR/IBR always go down once the matches redundancy is taken into
    account.
    %
    %
}
        \label{fig:matching:detections}
\end{figure*}

\section{Discussion}\label{sec:concl}

The observation that the classic repeatability criterion does not take spatial redundancy into account has motivated the
introduction of a performance metric: the non-redundant repeatability. It is an
adaptation of the classic criterion involving the region
covered by the descriptor.
To illustrate the new repeatability criterion, the performance of several
state-of-the-art methods were examined. We observed that, once the descriptors
overlap is taken into account, the traditional hierarchy of the methods was severely
disrupted. The detections and associated description generated by some methods
are highly correlated.
Such redundant parasite detections are arguably caused by scale-space sampling issues (as in
the case of Hessian and Harris based methods)
or the method's design. For example, the SIFER's kernel generates clusters of scale space extrema for each blob.
A reassuring characteristic of the new repeatability criterion is that it seems to be
in agreement with the redundancies observed on patterns and on natural images and that it also agrees with the matching performance when using the same description technique for all methods.
Overall, the SIFT and SFOP methods appear to perform best as they offer the best balance between a large number of detections and a strong non-overlapping repeatability.
SIFT and SFOP detections also seem to be complementary, each one
detecting different image features.
Regarding the non-redundant repeatability the variant SIFT-single
beats SIFT on all the analyzed sequences, and therefore seems
to be a recommendable replacement for SIFT.
For most benchmark data and particularly for those with strong affine distortion,  MSER  performs best in non-redundant repeatability.

\section*{Acknowledgements}
Work  partially supported by  Centre National d'Etudes Spatiales
(CNES, MISS Project),  European Research Council (Advanced Grant Twelve
Labours),  Office of Naval Research (Grant N00014-97-1-0839), Direction
G\'{e}n\'{e}rale de l'Armement (DGA), Fondation Math\'{e}matique Jacques
Hadamard and Agence Nationale de la Recherche (Stereo project).

\bibliographystyle{IEEEtran}
\bibliography{main_v3.2_arxiv}

\end{document}